\documentclass[twocolumn]{article}          % twocolumn
\pdfoutput=1
\usepackage{mathptmx}      % use Times fonts if available on your TeX system
\usepackage[margin=1in]{geometry}
\usepackage{times}
\usepackage{epsfig}
\usepackage{epstopdf}
\usepackage{graphicx}
\usepackage{amssymb}
\usepackage{amsmath}
\usepackage{latexsym}
\usepackage{multicol}
\usepackage{booktabs}
\usepackage{mathtools}
\usepackage{multirow}
\usepackage{wrapfig}
\usepackage{ulem} % check SMH if can be used %
\usepackage{algorithm}
\usepackage{algorithmic}
\usepackage{enumerate}
\usepackage{color}
\usepackage{tikz}
\usepackage{subfig}
\usepackage{multirow}
\usepackage{array}
\usepackage{adjustbox}
\usepackage[numbers]{natbib}
\usepackage{titlesec}
\usepackage{authblk}

\titleformat{\section}
  {\normalfont\fontsize{10}{12}\bfseries}{\thesection}{1em}{}
\titleformat{\subsection}
  {\normalfont\fontsize{10}{12}\bfseries}{\thesubsection}{1em}{}
\titleformat{\subsubsection}
  {\normalfont\fontsize{10}{12}\bfseries}{\thesubsubsection}{1em}{}

  \usetikzlibrary{shapes,arrows, trees}

\newcommand\fnote[1]{\captionsetup{font=small}\caption*{#1}}

\begin{document}

\title{A Face Fairness Framework for 3D Meshes}

\author{Sk. Mohammadul Haque}
\author{Venu Madhav Govindu}
\affil{Indian Institute of Science, Bengaluru}
\date{}
\maketitle

\begin{abstract}
In this paper, we present a face fairness framework for 3D meshes that preserves the regular shape of faces and is applicable to a variety of 3D mesh restoration tasks. Specifically, we present a number of desirable properties for any mesh restoration method and show that our framework satisfies them. We then apply our framework to two different tasks --- mesh-denoising and mesh-refinement, and present comparative results for these two tasks showing improvement over other relevant methods in the literature.
\end{abstract}

\section{Introduction\label{sec:Introduction}}

Although in recent years it has become easier to acquire 3D data using either depth cameras or by solving a dense multi-view stereo problem using RGB images, the inherent limitation of the quality of raw 3D data available is still present. For example, depth representations obtained from depth cameras contain significant amount of noise both due to the quantisation and also due to the estimation technique involved. In other acquisition modalities like multi-view stereo which generate point clouds, isotropic noise is expected to be present. Thus meshes obtained from such modalities contain noise that are distributed in all the directions and all 3D reconstruction pipelines that deal with such meshes require a denoising scheme. However, even after denoising, meshes that are obtained either directly from depth cameras or from multi-view stereo are often of poor quality in terms of details. Recently, there has been significant research towards improving the quality of these meshes through explicit photometric information \cite{Wu2011Fusing, Haque2014High, Kim2016MVIR, Haque2017MVRefinement}. 
In almost all the 3D reconstruction pipelines that generate meshes, it is desired that the quality of the final meshes obtained is also adequate in terms of a certain number of properties like smoothness while simultaneously preserving details, quality of the face shapes, \textsl{etc.} 

In this paper, a general face fairness framework for improving the quality of faces in 3D meshes is proposed. This is an extended version of our work in \cite{Haque2015Mesh}. Our proposed  encourages regular flat faces thereby preventing skinny and flipped faces and ensures smoothness while simultaneously preserving surface details. Our framework is very general and is applicable to a variety of 3D mesh processing or enhancing tasks.\\

\subsection{Related work}
Fairness of face shapes deals with the quality of the shape of the faces in 3D meshes. Early works include those from finite element methods. In these methods, approximate solutions to partial differential equations are computed in the form of piecewise linear functions using triangle meshes. Theoretical conditions have been developed that determine good shapes of the triangles such that the errors in the approximations are minimised \cite{Fried1960Condition, Babuvska1976Angle}. A popular criterion is the Delaunay triangulation that guarantees a number of geometric aspects of well-shaped triangles \cite{Du1995Computing}. Unlike these works where the mesh topology is either created or improved, in many situations the mesh topology is retained and only the vertex positions are modified. Such examples are abundant in 3D reconstruction pipelines.
\textsl{e.g.} 3D mesh denoising, 3D mesh refinement.

For denoising, 3D reconstruction pipelines either explicitly apply a denoising algorithm on the input data or implicitly use a smoothing prior during processing of the raw 3D data.
There are many methods for explicit 3D mesh denoising proposed in the literature. We may classify most of these approaches into a) local, or b) global methods. In \textit{local} methods, the correction for the noisy mesh is applied locally, resulting in approaches that are iterative in nature~\cite{Field1988Laplacian,Fleishman2003Bilateral,Sun2007FastEffective,Sun2008RandomWalk,Taubin2001LinearAniso,Zheng2011Normal}. Often simple local methods also result in artefacts like geometric distortion and surface shrinkage~\cite{Liu2007NonIterative,Taubin2001LinearAniso}. In \textit{global} methods, a global cost function is optimised~\cite{Desbrun1999Implicit,Hoppe1993Optimization,Ji2005Global,Liu2007NonIterative,Nealen2006Laplacian,Ohtake02MeshSmoothing,Zheng2011Normal}. This typically involved solving a sparse system of equations that are usually linear in nature~\cite{Liu2007NonIterative}. More recent methods of removing noise from meshes are engineered to recover as much detail as possible \cite{Wei2015BiNormal, Lu2017Efficient, Lu2017RobustPre}.

On the other hand, the implicit smoothing priors are very frequently used in methods that use an optimisation framework in contexts other than explicit denoising. For example Wu \textit{et al.}~\cite{Wu2011HQMultiviewStereo} uses an anisotropic smoothing prior in their shading based mesh-refinement step to avoid noisy reconstructions in areas that are ill-constrained. However, methods that do not use such smoothing regularisers ~\cite{Nehab2005Combine,Wu2011Fusing} suffer from irregular faces and often noisy artefacts in the final reconstructed meshes. Ideally, these methods assume that the initialisation is smooth and very near to their optimal solution. Very different from directly defining smoothing priors on meshes, methods like TSDF~\cite{Curless1996Volumetric}, VolumeDeform~\cite{Innmann2016Volumedeform} define smoothing priors across 3D voxels.

% ####EDIT HERE
\begin{figure}[h]
\centering\includegraphics[width=0.9\linewidth]{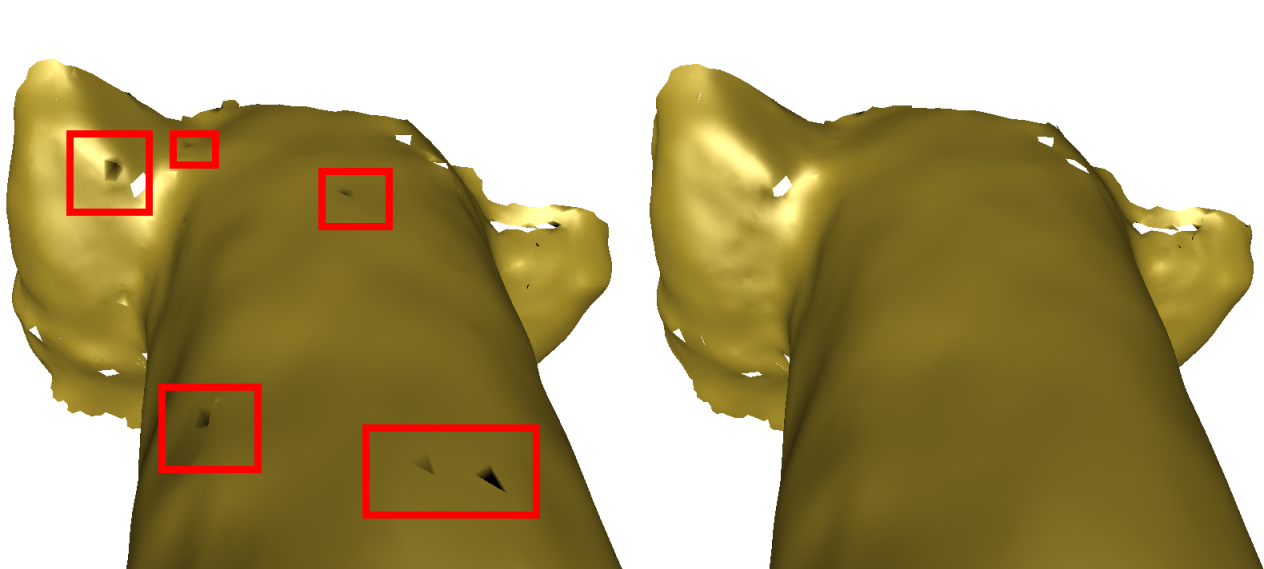}\\
\protect\caption{Flipped faces of a denoised mesh. Left image: Sun \textsl{et al.}~\cite{Sun2007FastEffective}. Right image: Desired result.} \label{fig:Fair_comparison}
\end{figure}

\begin{figure}[h]
\centering\includegraphics[width=0.31\linewidth]{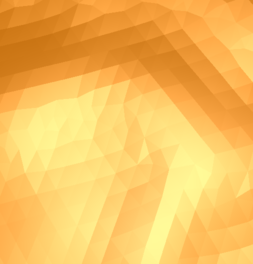}\,\,\includegraphics[width=0.31\linewidth]{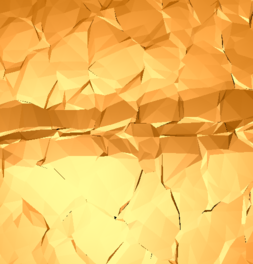}\,\,\includegraphics[width=0.31\linewidth]{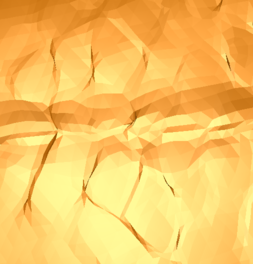}\\
\protect\caption{Surface breakage of a mesh during photometric refinement using a naive approach of \cite{Sun2007FastEffective}. The columns correspond to initial smooth mesh, refined output from \cite{Sun2007FastEffective} and output of 1 iteration of Laplacian smoothing on the refined mesh respectively.} \label{fig:Surface_breakage}
\end{figure}

Many of these mesh denoising algorithms and smoothing prior frameworks ignore the true distribution of observation noise in depth representations and assume the noise to be restricted along the direction of the surface normal. As a result, they correct for the position of a mesh vertex (point) by moving it along this normal direction. The noise component in the tangent plane about a surface point is completely ignored. For mesh denoising, this works well for low noise scenarios and is reasonable from the surface recovery perspective~\cite{Fleishman2003Bilateral}. But at higher noise levels, the noise component in the tangent plane leads to a severe distortion of the face shapes in the mesh, including face flipping with the surface normals being forced to point into a surface rather than out of it. Figure \ref{fig:Fair_comparison} illustrates such a scenario when denoising a noisy 3D mesh of an archaeological structure. The left image shows that the denoising using the method of Sun \textsl{et al.}~\cite{Sun2007FastEffective} introduces flipped faces that are rendered as black patches in smooth-shaded rendering mode. However, ideally we desire to obtain a denoised mesh as the right image.  In mesh refinement frameworks where surface-details are fused into an initial smooth mesh lacking details, like ~\cite{Nehab2005Combine,Wu2011Fusing}, the vertices are allowed to move only along the normal directions. Such restriction with fewer degrees of freedom does not allow the mesh to fit the fine scale details adequately and instead undesired artefacts like surface breakage develop during the refinement process. For such surfaces, subsequent post-processing completely fails to remove such artefacts. For example, Figure \ref{fig:Surface_breakage} shows the application of  mesh-refinement technique of \cite{Sun2007FastEffective} on a portion of an initial smooth mesh (left) with a high quality normal map obtained through photometric stereo. The centre image shows the introduction of the surface-breakage artefacts after application of \cite{Sun2007FastEffective}. An attempt to remove the artefacts using a mild Laplacian filtering aggravates the situation by smoothing the details. The reason for such artefacts is that the initial mesh is too smooth and is very far from the final expected mesh and the movement of vertices only along the normal directions is inadequate to solve the task. 

Apart from carefully accounting for the presence of noise in different directions, mesh  frameworks including denoising methods should also avoid typical problems such as volume shrinkage and smoothing over surface features such as edges and corners. Although the classical Laplacian mesh smoothing method~\cite{Field1988Laplacian} does account for noise in all directions, methods based on such Laplacian smoothing do not preserve surface features and are also affected by the problem of volume shrinkage~\cite{Sun2007FastEffective,Taubin2001LinearAniso}. More recent works that present methods designed to preserve features include~\cite{Cheng2014FeatureL0,Fan2010Robust,He2013MeshL0,Ji2005Global,Jones2003NonIterative,Sun2007FastEffective,Sun2008RandomWalk, Wang2014DNF,Zhang2015Variational,Zheng2011Normal}. The methods by He \textsl{et al.}~\cite{He2013MeshL0} and Cheng \textit{et al.}~\cite{Cheng2014FeatureL0} are good for piecewise flat surfaces. However, they result in introducing artificial edges in smooth regions and hence work poorly on natural data. Moreover some methods either involve complicated stages or expensive optimisations. For example, the method of He \textsl{et al.}~\cite{He2013MeshL0} solves a sequence of expensive minimisation tasks to obtain the final results.\\

Given such a context of 3D mesh optimisation, we now state the issues that a mesh optimisation framework should address as follows:

\begin{itemize}
\item \textbf{Feature-preserving smoothing} - The final mesh should be as smooth as possible while preserving all the surface details. This requires an explicit smoothing prior in the cost function.
\item \textbf{Degrees of freedom} -  A framework should allow enough degrees of freedom to the optimising vertices to fit the assumed model sufficiently.
\item \textbf{Face fairness} - The shapes of the faces in the optimised mesh should be prevented from becoming skinny and getting folded.
\item \textbf{Generalisation} - The framework should be general enough to be applicable to a variety of tasks.
\item \textbf{Efficiency} - The framework should be efficient to be run on large meshes.
\end{itemize}

In Section~\ref{sec:smoothframework}, we describe our face fairness framework for 3D meshes. Specifically, in Section~\ref{subsec:Definitions-and-notations} we briefly describe the observation model for a 3D mesh and give a general perspective of a class of approaches to how mesh vertices are optimised. Section \ref{subsec:framework} lays down our fairness framework for meshes that is based on minimising a global cost function that is quadratic and sparse in nature and explicitly incorporates a measure of face fairness. After that we demonstrate the application of our framework in two different contexts, namely, mesh denoising in Section~\ref{subsec:Denoising} and mesh normal fusion in Section~\ref{subsec:mesh-normal-fusion}. We also present extensive results of our method on a variety of datasets and compare our performance with that of other related approaches in the literature.\\

\section{Proposed face fairness framework} \label{sec:smoothframework}
\subsection{Preliminaries and mesh degradation} \label{subsec:Definitions-and-notations}

We consider a clean oriented surface $\mathbf{S}_{0}$ which is piece-wise smooth. Let $\mathbf{M}_{0}$ be an oriented mesh approximated from $\mathbf{S}_{0}$ and is given as 
 \begin{equation}
\mathbf{M}_{0}=\left(\mathbf{V}_{0},\mathbf{E}_{0},\mathbf{F}_{0}\right).
 \end{equation}
Here $\mathbf{V}_{0}=\left\{ \mathbf{v}_{0i}\right\} _{i=1}^{N_{V}}$ is the noise-free set of the $N_{V}$ sampled points in $\mathbf{S}_{0}$. $\mathbf{E}_{0}$ is the set of $N_{E}$ 1D edges given as 
\begin{equation}
\mathbf{E}_{0}=\left\{ \mathbf{e}_{i}|\mathbf{e}_{i}=\left(\mathbf{v}_{p},\mathbf{v}_{q}\right)\in\mathbf{V}_{0}\times\mathbf{V}_{0}\right\} _{i=1}^{N_{E}}
\end{equation}
and $\mathbf{F}_{0}$ is the set of $N_{F}$ triplets defining the 2D faces of the mesh and is given as
\begin{equation}
\begin{split}
\mathbf{F}_{0} = &\{ \mathbf{f}_{i}|\mathbf{f}_{i}=\left(\mathbf{v}_{p},\mathbf{v}_{q},\mathbf{v}_{r}\right)\in\mathbf{V}_{0}\times\mathbf{V}_{0}\times\mathbf{V}_{0}, \\
&   \qquad \left(\mathbf{v}_{p},\mathbf{v}_{q}\right), \left(\mathbf{v}_{q},\mathbf{v}_{r}\right),\left(\mathbf{v}_{r},\mathbf{v}_{p}\right)\in\mathbf{E}_{0}\} _{i=1}^{N_{F}}.
\end{split}
\end{equation}
We note here that for convenience we will use the term $\mathbf{v}_i$ for both the $i^{th}$ vertex itself and its position and the meaning can be easily identified from the context. Similarly, based on the context, we will use the term $\mathbf{V}$ for both the set of vertices itself and the 1D concatenation of their positions ordered by their indices. Also, since we consider only triangle meshes, we will use the terms triangle and face interchangeably.

\begin{figure}[h!]
\noindent \begin{center}
\includegraphics[width=0.5\linewidth]{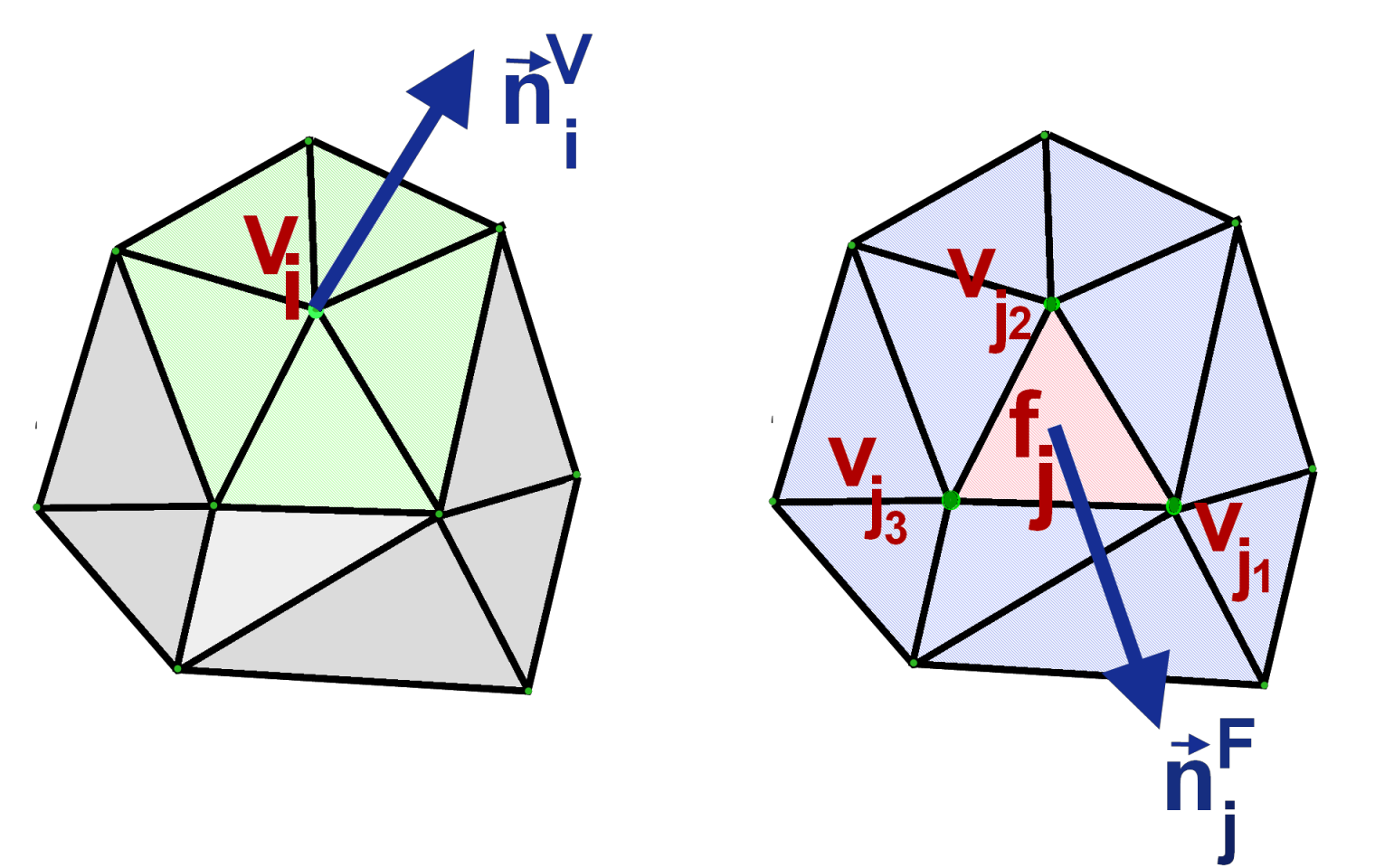}
\par\end{center}
\protect\caption{Conventions used in the paper: The left image shows a portion of a mesh where $\mathbf{v}_i$ is a vertex with a normal $\mathbf{n}_i^{V}$. The green shaded area around $\mathbf{v}_i$ is the face neighbourhood of $\mathbf{v}_i$ used in the paper. The right image represents the same portion of a mesh where $\mathbf{f}_j$ is a face with vertices $\mathbf{v}_{j_{1}}, \mathbf{v}_{j_{2}}, \mathbf{v}_{j_{3}}$. This face has a normal $\mathbf{n}_j^{F}$ defined at its centroid. The blue shaded area consisting of all the faces around $\mathbf{f}_j$ is the face neighbourhood used in the paper.\label{fig:Notations}}
\end{figure}

As shown in Figure \ref{fig:Notations}, the left image shows a normal $\mathbf{n}_i^{V}$ defined at a vertex $\mathbf{v}_i$ which is usually computed using the normals defined on the neighbouring faces shown in green shade. For a face $\mathbf{f}_{j}$ which is defined by the vertices $\mathbf{v}_{j_{1}}, \mathbf{v}_{j_{2}}, \mathbf{v}_{j_{3}}$ in counter-clockwise order, the normal vector $\mathbf{n}_{j}^{F}$ at its centroid is same as the normal of the plane in which the triangle resides. A number of averaging schemes are used for the computation of the vertex normal. In the simplest cases, the mean of the surrounding faces is taken. However, it is consistent with the geometry only in very smooth regions. To compute vertex normals more accurately, weighted averaging schemes based on incident angles \cite{Thurrner1998Computing}, incident edge lengths \cite{Max1999Weights}, \textsl{etc.} considering the discrete mesh topology are proposed in the literature. A comprehensive discussion can be found in \cite{Meyer2003Discrete}.

 The boundary of a mesh $\mathbf{M}$ is denoted as an ordered pair $\partial\mathbf{M}=\left(\mathbf{V}^{B},\mathbf{E}^{B}\right)$ where $\mathbf{V}^{B}\subset\mathbf{V}$, $\mathbf{E}^{B}=\left\{ \left(\mathbf{v}_{p},\mathbf{v}_{q}\right)\right\} \subset\mathbf{E}$ and $\mathbf{v}_{p},\mathbf{v}_{q}\in\mathbf{V}^{B}$. By $\mathcal{N}\left(\mathbf{x}\right)$, we denote a neighbourhood operator on an entity $\mathbf{x}$ which can be either a vertex or a face, depending on the context.

\textit{\textbf{Observation Model}}: A general model of degraded observation for any vertex $\mathbf{v}_{i}$ can be given as
 \begin{equation}
\mathbf{v}_{i}=\mathbf{v}_{i0}+ l_i\mathbf{s}_{i} \label{eqn:observation_model}
 \end{equation} 
 where $\mathbf{v}_{i0}\in\mathbf{V}_{0}$ is the true noise-free but unknown vertex set, $\mathbf{s}_{i}$ is the additive degradation usually assumed as zero-mean $\sigma^2$-variance \textit{i.i.d.} Gaussian noise and $l_i$ is the local scale of sampling around vertex $i$. We assume that the observed topology i.e. the face set and the edge set are unaltered during the generative process and focus on the noise model on the vertex set. Hence, the observed mesh can be represented as $\mathbf{M}=\left(\mathbf{V},\mathbf{E}_{0},\mathbf{F}_{0}\right)$ where $\mathbf{V}=\left\{ \mathbf{v}_{i}\right\} _{i=1}^{N_{V}}$ is the degraded set of vertex position measurements. \\

\textit{\textbf{Vertex Modification}}: Before proceeding further, we would like to mention a commonly used general form of vertex modification~\cite{Taubin2001LinearAniso}. It is given as 
\begin{equation}
\widehat{\mathbf{v}}_{i}=\mathbf{v}_{i}+{\displaystyle \sum_{j}}w_{ij}\mathbf{A}_{ij}\mathbf{v}_{j} \label{eqn:vertex_modification_step}
\end{equation}
where $\widehat{\mathbf{v}}_{i}$ is the estimated vertex, $\mathbf{A}_{ij}$ is a linear operator defined locally around $\mathbf{v}_{i}$ and $w_{ij}$ are the weights on the corresponding neighbouring vertices $\mathbf{v}_{j}\in\{\mathcal{N}\left(\mathbf{v}_{i}\right)\cup i \}$. Such a vertex modification is often iteratively applied~\cite{Fleishman2003Bilateral,ElOuafdi2008PhysicalManifold,Taubin1995Signal}. The weights $w_{ij}$ and the local operator $\mathbf{A}_{ij}$ vary depending on the algorithm. In Laplacian smoothing~\cite{Field1988Laplacian}, $\mathbf{A}_{ij}=\mathbf{I}$, \textsl{i.e.} the identity operator which is isotropic in nature, weights $w_{ij}$ can be either constant or depend on the corresponding face areas or cotangents. Such Laplacian smoothing does not preserve features and also results in high volume shrinkage. To mitigate these problems to an extent, non-identity $\mathbf{A}_{ij}$'s have been used, \textsl{e.g.} bilateral mesh filtering~\cite{Fleishman2003Bilateral} where $w_{ij}$ are set to the bilateral weights and $\mathbf{A}_{ij}$ are the orthogonal projections onto the respective normal directions resulting in anisotropy.

 Many methods like \cite{Nehab2005Combine, Wu2011Fusing, Kim2016MVIR} use an optimisation framework to minimise a global energy. Interestingly, these methods implicitly use  the vertex improvement iterations of Equation \ref{eqn:vertex_modification_step}. They move the vertices only along their normal directions to avoid self-intersections leading to a final update step as

\begin{equation}\label{key}
\widehat{\mathbf{v}}_{i}=\mathbf{v}_{i}+\delta_{i}\mathbf{n}_{i}
\end{equation}
 where $\delta_{i}$ is obtained from optimising their cost function. Such a restriction of movement of vertices to their normal directions is often insufficient in tasks like mesh refinement. In \cite{Nehab2005Combine, Wu2011Fusing}, no explicit smoothing prior is used. As we will show later in Section \ref{subsec:mesh-normal-fusion}, such methods suffer from presence of noise in their output meshes. 

\subsection{Our framework\label{subsec:framework}}

Collecting the set of $N_{V}$ vertices $\left\{ \widehat{\mathbf{v}}_{i}\right\} _{i=1}^{N_{V}}$ into a single concatenated vector denoted as $\mathbf{V}$, we denote the initial observed vertices as $\mathbf{V}$ and the vertices obtained after applying our optimisation framework as $\widehat{\mathbf{V}}$ respectively. In our mesh optimisation framework, we minimise a global cost function $C_{V}(\widehat{\mathbf{V}})$  which has three terms, \textit{namely},
\begin{enumerate}
	\item a data observation term $d_{o}^{V}(\widehat{\mathbf{V}},\mathbf{V})$,
	\item a Laplacian smoothing term $d_{s}^{V}(\widehat{\mathbf{V}})$, and
	\item a face fairness term $d_{f}^{V}(\widehat{\mathbf{V}})$.
\end{enumerate}
    The resulting cost function to be minimised becomes 
\begin{equation}
C_{V}(\widehat{\mathbf{V}}) = d_{o}^{V}(\widehat{\mathbf{V}},\mathbf{V}) +\lambda_{V} d_{s}^{V}(\widehat{\mathbf{V}}) + \eta d_{f}^{V}(\widehat{\mathbf{V}})\label{eq:vertex_updation}
\end{equation}

where $\lambda_{V}$ are $\eta$ are parameters depending only on the type and amount of noise.
\subsubsection{Data term}
From our observation model in Equation \ref{eqn:observation_model} for each vertex $\mathbf{v}_i$, we have that the quantity $\mathbf{u}_i$ given as
\begin{equation}
\mathbf{u}_i = \dfrac{\mathbf{v}_{i0}-\mathbf{v}_i}{l_i}
\end{equation}
 has a normalised \textit{i.i.d.} 3D Gaussian distribution. Hence, the data term is a quadratic penalty given as 
 \begin{equation}
 d_{o}^{V}\left(\widehat{\mathbf{V}},\mathbf{V}\right)={\displaystyle \left\Vert \mathbf{W}\left(\widehat{\mathbf{V}}-\mathbf{V}\right)\right\Vert _{2}^{2}}
 \end{equation}
 where $\mathbf{W}$ is a diagonal matrix with entries $w_{ii}=\tfrac{1}{l_{i}}$. Since the true local scale $l_{i}$ is unknown, it is typically estimated from the local neighbourhood around $\mathbf{v}_i$ in the degraded mesh $\mathbf{M}$ itself. However, we will assume that the scale does not vary significantly and hence $\mathbf{W}=\mathbf{I}$ is an identity matrix.
  
\subsubsection{Our Laplacian smoothing term}

Our Laplacian operator is anisotropic in nature which is defined only along the normal directions at the vertices. However, in our approach we have carefully selected a bilateral weighting scheme. Specifically we use the following operator $\mathbf{L}_{i}\left(\cdot\right)$ to define the Laplacian form for each vertex $\mathbf{v}_{i}$ \textsl{i.e.}

\begin{equation}
\mathbf{L}_{i}\left(\mathbf{v}_{i}\right) =\displaystyle \sum_{j\in\mathcal{N}_{V}\left(i\right)} w_{ij} \mathbf{A}_{j} \left(\mathbf{v}_{i} - \dfrac{\left(\mathbf{v}_{j_{1}}+\mathbf{v}_{j_{2}}+\mathbf{v}_{j_{3}}\right)}{3}\right) \label{eqn:laplacian}
\end{equation}

where 
\begin{equation}
 w_{ij} = \left(\dfrac{a_{ij}b_{ij}}{\left(1+a_{ij}\right){\displaystyle \sum_{j\in\mathcal{N}_{V}\left(i\right)}b_{ij}}}\right)
\end{equation}
and 
\begin{equation}
\mathbf{A}_{j} = \left(\mathbf{n}_{j}^{F}\mathbf{n}_{j}^{F,T}\right). \label{eqn:A_operator}
\end{equation}
Here, $a_{ij}$ and $b_{ij}$ form the bilateral weighting functions and $\mathbf{n}_{j}^{F}$ is the normal of the neighbouring face $\mathbf{f}_{j}$ corresponding to the vertex $\mathbf{v}_{i}$ \textsl{i.e.} $\mathcal{N}_{V}\left(i\right)$ is the set of 1-ring neighbouring faces to the vertex $\mathbf{v}_{i}$ and $\mathbf{v}_{j_{1}}$, $\mathbf{v}_{j_{2}}$, $\mathbf{v}_{j_{3}}$ are vertices of $\mathbf{f}_{j}$. We denote $\Delta{\mathbf{v}_{ij}}$ as the difference between the vertex $\mathbf{v}_{i}$ and the centroid of face $\mathbf{f}_{j}$, \textsl{i.e.} $\Delta{\mathbf{v}_{ij}} = \frac{\mathbf{v}_{j_{1}} + \mathbf{v}_{j_{2}} + \mathbf{v}_{j_{3}}}{3} - \mathbf{v}_{i}$. Consequently, for our approach we define 

\begin{equation}
a_{ij}=\exp \left(-\dfrac{\left( \mathbf{n}_{j}^{T}\Delta{\mathbf{v}_{ij}}\right)^{2}}{2 \sigma_{1}^{2}\l_{i}^{2}}\right)
\end{equation}

which is the weighting along the normal direction of the considered neighbouring face corresponding to the vertex $\mathbf{v}_{j}$. Similarly, we have 
\begin{equation}
b_{ij}=\exp\left(-\dfrac{\left\Vert \Delta{\mathbf{v}_{ij}} \right\Vert _{2}^{2}}{2 \sigma_{2}^{2}\l_{i}^{2}}\right).
\end{equation}

We note here that each Laplacian $\mathbf{L}_{i}\left(\mathbf{v}_{i}\right)$ is a function pertaining to a single vertex and all such terms are concatenated to form our global weighted Laplacian $\mathbf{L}\left(\mathbf{V}\right)$ used in Equation~\ref{eq:vertex_updation}.

\subsubsection{Face fairness penalty} \label{subsubsec:face_fairness}
\begin{figure}
\noindent \begin{center}
\includegraphics[width=0.5\linewidth]{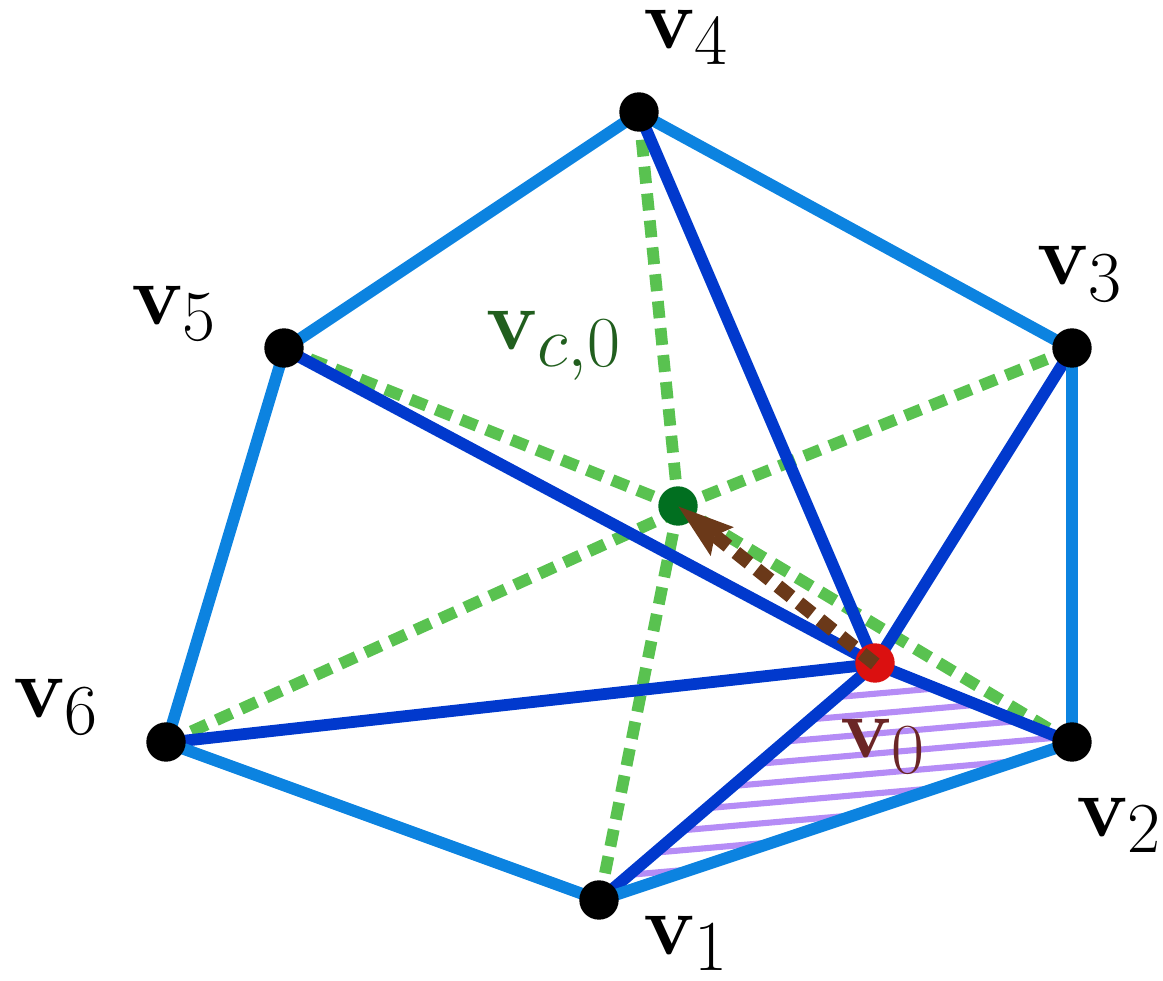}
\par\end{center}

\protect\caption{Illustration of our face fairness penalty} \label{fig:face_demons}
\end{figure}
As discussed in previous sections, while most optimisation methods apply vertex correction only along the surface normal, the actual noise or imperfection present in a mesh also has a component that lies in the tangent plane about a vertex. Neglecting this fact even under moderate noise levels leads to irregular-shaped faces and very often undesirable folding of faces. In our mesh optimisation framework, we desire that the shape of the faces in the estimated mesh become as regular as possible without changing its geometry. To integrate this into our framework, we explicitly introduce a face fairness term $d^{V}_{f}$ that ensures triangular faces do not become skinny or folded. The face fairness penalty for a single estimated vertex  $\widehat{\mathbf{v}}_{i}$, is

\begin{equation}
d_{f}^{V}\left(\widehat{\mathbf{v}}_{i}\right)=\left\Vert r_{i}(\mathbf{I}-\mathbf{n}_{i}^{V}\mathbf{n}_{i}^{V,T})(\widehat{\mathbf{v}}_{c,i}-\widehat{\mathbf{v}}_{i})\right\Vert_2^2 \label{eqn:fairness-function-per-vertex}
\end{equation}
where $\widehat{\mathbf{v}}_{c,i}$ is the centroid of the 1-ring face neighbourhood $\mathcal{N}_{V}\left(i\right)$ around the vertex $\mathbf{v}_{i}$ and the weight $r_{i}$ is given as

\begin{equation}
r_{i}=\begin{cases}
0 & \textrm{if } \mathbf{v}_{i}\in\mathbf{V}^{B}\\
\beta & \textrm{otherwise}
\end{cases}\label{eq:fairness_weight}
\end{equation}

where $\beta = \textrm{max} \{\displaystyle\textrm{mean}_{p,q\in\mathcal{N}_{V}\left(i\right)}(\mathbf{n}_{p}^{F,T}\mathbf{n}_{q}^{F}-\delta),0\}$ and $\delta$ is a small positive value ($\delta\sim0.2$). Note that that in \cite{Haque2015Mesh}, $\beta$ was chosen as $\displaystyle\min_{p,q\in\mathcal{N}_{V}\left(i\right)}(\mathbf{n}_{p}^{F,T}\mathbf{n}_{q}^{F}-\delta)$. However, this was susceptible to erroneous normals. Choosing the mean value instead of minimum increases its robustness. Also, we note that the first condition in the weight defined in Equation~\ref{eq:fairness_weight} ignores the boundary in open meshes whereas the second condition carefully gives less weight to the edges and corners. Also, the fairness penalty constrains the solution only in the tangential plane about a vertex without affecting the Laplacian smoothness term significantly.
\begin{figure}[h]
\noindent \begin{center}
\includegraphics[width=0.85\linewidth]{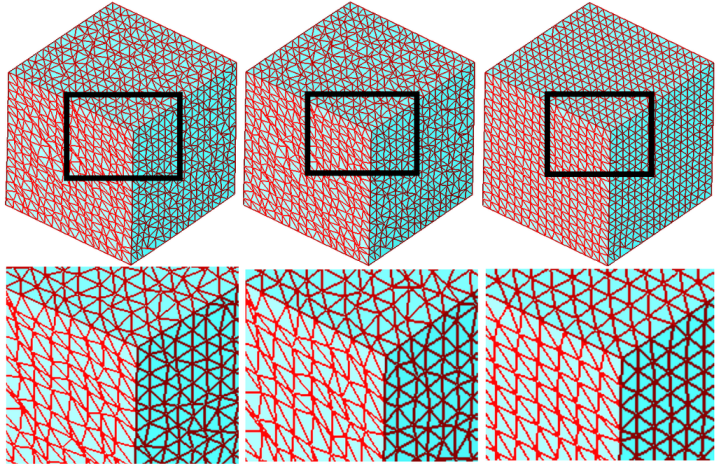}
\par\end{center}

\protect\caption{Face quality of denoised mesh of a cube ($N_V=1538,N_F=3072$) corrupted with isotropic Gaussian noise with standard deviation $\sigma = 0.15\times \textrm{mean edge length}$.  Left image: Sun \textsl{et al.}~\cite{Sun2007FastEffective}. Middle image: Our method without fairness penalty. Right image: Our method with fairness penalty. Our method with the fairness penalty is able to ensure face fairness whereas the other two methods fail to do so.} \label{fig:Fair_comparison_detail}
\end{figure}

\begin{table}
\noindent \begin{center}
{\footnotesize
\begin{tabular}{|c|c|c|c|}
\hline 
Error metric & Sun \cite{Sun2007FastEffective} & \begin{minipage}{0.75in}Ours \\(w/o fairness term $d_{f}^{V}$) \end{minipage} & \begin{minipage}{0.75in}Ours\\ (with fairness term $d_{f}^{V}$) \end{minipage} \tabularnewline
\hline
\hline
Mean NE ($^{\circ}$) & 0.6427 & 0.5704 & 0.4633 \tabularnewline
\hline
Mean VPE & 0.0259 & 0.0258 & 0.0129 \tabularnewline
\hline 
\end{tabular}
}
\par\end{center}
\protect\caption{Comparison of errors in the denoised output for different methods on the cube example given in Fig.~\ref{fig:Fair_comparison_detail}. NE denotes `normal angle error' and VPE denotes `vertex position Euclidean distance error'.}\label{table:cube_quality}
\end{table}

Figure \ref{fig:face_demons} explains how our face fairness penalty encourages well-shaped triangles. We consider a 1-ring neighbourhood around the vertex $\mathbf{v}_0$ in a portion of a mesh that has poorly-shaped faces. The neighbourhood contains the seven vertices $\mathbf{v}_i$, $i=0,1,\cdots,6$ and the six triangles formed by them. Clearly, the shaded face $\mathbf{f}_0$ formed by $\mathbf{v}_0$, $\mathbf{v}_1$ and $\mathbf{v}_2$ is skewed in shape. From Equation \ref{eqn:fairness-function-per-vertex}, for vertex $\mathbf{v}_0$, we have

\begin{equation}
 d_{f}^{V}\left(\mathbf{v}_{0}\right)=\left\Vert r_{0}(\mathbf{I}-\mathbf{n}_{0}^{V}\mathbf{n}_{0}^{V,T})(\mathbf{v}_{c,0}-\mathbf{v}_{0})\right\Vert_2^2. \label{eqn:sample_equation_per_vertex}
\end{equation}

The factor $(\mathbf{I}-\mathbf{n}_{0}^{V}\mathbf{n}_{0}^{V,T})(\mathbf{v}_{c,0}-\mathbf{v}_{0})$ measures the projection of the difference of between $\mathbf{v}_{c,0}$ and $\mathbf{v}_0$ onto tangent space at $\mathbf{v}_0$. The factor $r_0$ weights the distance according to the amount of flatness at $\mathbf{v}_0$. Keeping everything else constant, minimisation of the cost in Equation \ref{eqn:sample_equation_per_vertex} over $\mathbf{v}_0$ leads to the movement of $\mathbf{v}_0$ towards $\mathbf{v}_{c,0}$ in the tangent plane and thereby minimises the disparity between the magnitude of the angles of $\mathbf{f}_0$.

The significance of our fairness penalty is illustrated in Fig.~\ref{fig:Fair_comparison_detail} where we compare the denoised mesh faces obtained by applying on an initial noisy cube, the method of Sun \textsl{et al.}~\cite{Sun2007FastEffective} and our method where we solve Equation~\ref{eq:vertex_updation} both with and without the fairness penalty term $d^{V}_{f}$. We can observe that the solutions of Sun \textsl{et al.}~\cite{Sun2007FastEffective} as well as our global method without the fairness penalty lacks the regular shape in the mesh faces. However, as is clearly evident, incorporating our fairness penalty term rectifies this problem and results in an accurate recovery of the mesh while also preserving the fairness of faces. In Table~\ref{table:cube_quality} we quantify the relative performance of the different methods in terms of the corresponding mean absolute errors of denoised normals in degrees and mean Euclidean error distances of the denoised vertices with respect to the ground truth. As can be seen, not only does the face fairness penalty improve the shapes of the denoised faces, it also improves the estimation of the vertex positions and the face normals.\\

\subsubsection{Optimisation}
We note here that since our cost function is quadratic, we have a closed form solution for our mesh fairness optimisation where the optimised $\widehat{\mathbf{V}}$ is given as 

\begin{equation}
\widehat{\mathbf{V}}=\left(\mathbf{I}+\lambda_{V}\mathbf{L}^{T}\mathbf{L}+\eta\mathbf{K}^{T}\mathbf{K}\right)^{-1}\left(\mathbf{V}\right) \label{eq:linear_solution_vertex}
\end{equation}

where $\mathbf{K}$ is formed from Equation \ref{eqn:fairness-function-per-vertex}. Since the cost function of Equation~\ref{eq:vertex_updation} is sparse in nature, we solve efficiently using gradient descent.

\subsubsection{Comparison with other face fairness penalties} \label{subsubsec:face_fairness_comp}
\begin{figure}[h]
\centering
\includegraphics[width=0.32\linewidth]{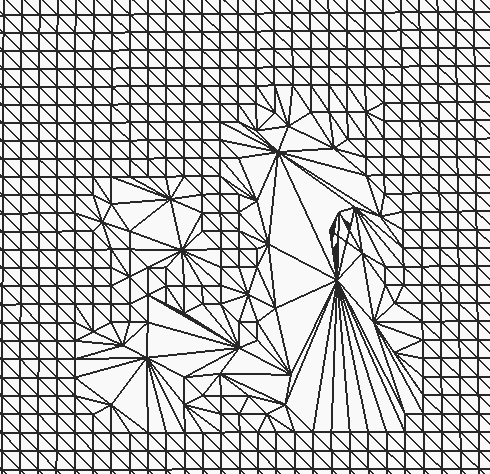}	\includegraphics[width=0.32\linewidth]{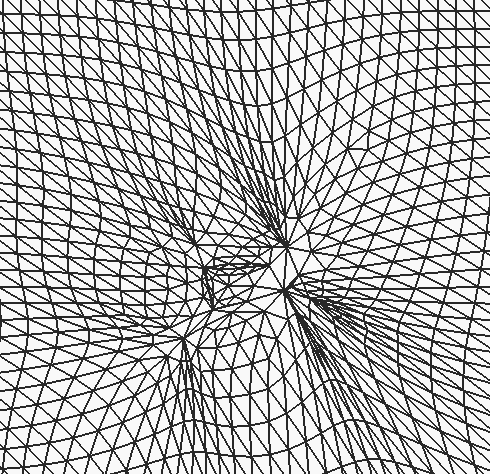}	\includegraphics[width=0.32\linewidth]{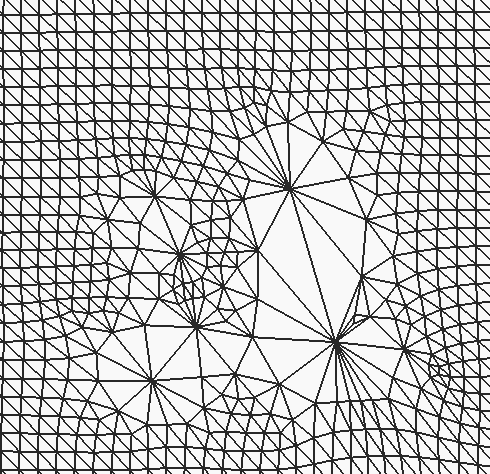}
\caption{Visual comparison of optimised meshes using method in \cite{Lu2017RobustPre} and our face-fairness. Left: Initial mesh. Centre: Optimised using method in \cite{Lu2017RobustPre}. Right: Optimised using our fairness.}	\label{fig:fair_plane}
\end{figure}

\begin{figure}[h]
\centering
\includegraphics[width=0.85\linewidth]{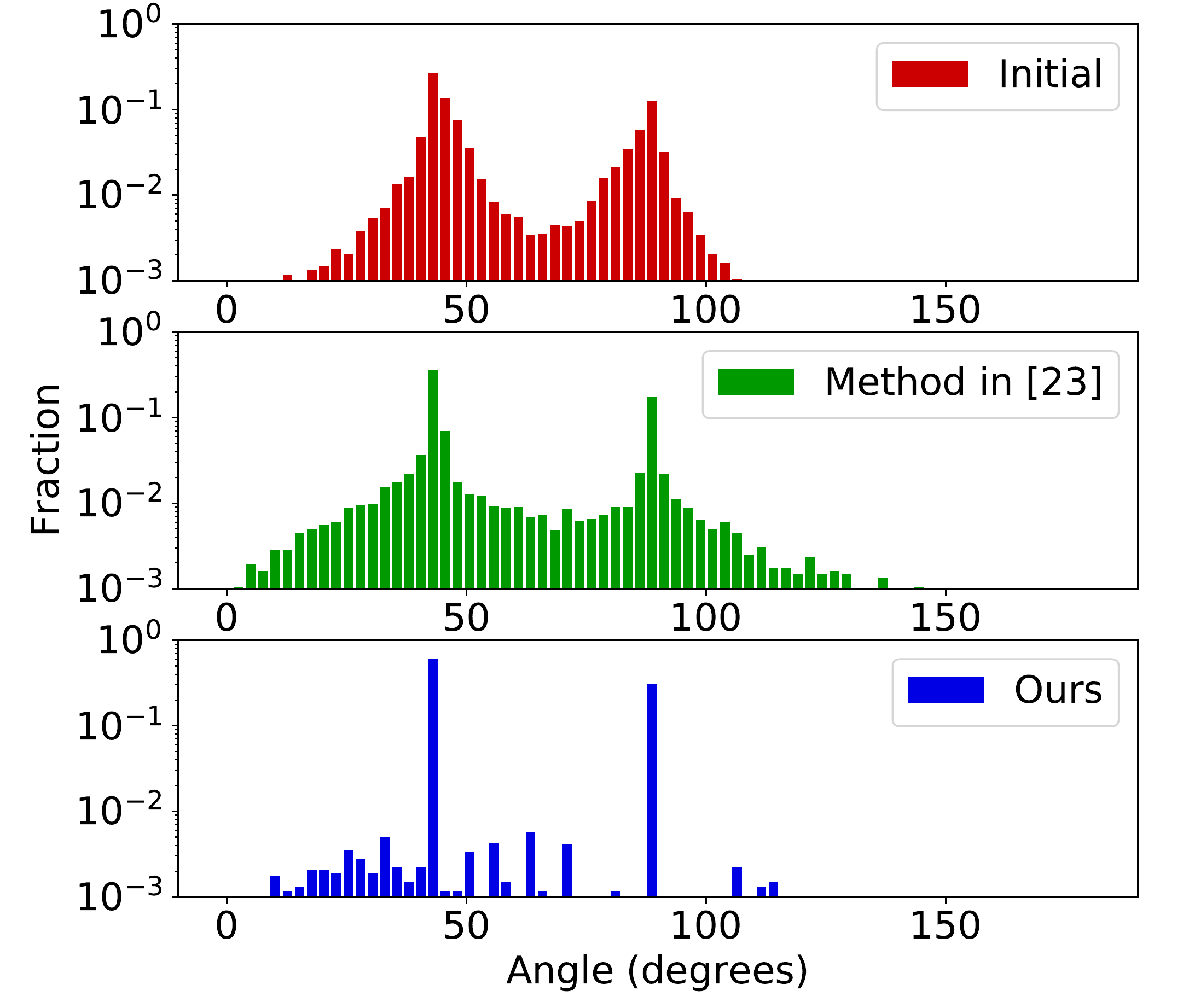}
\caption{Comparison of histograms of face corner angles in optimised meshes using method in \cite{Lu2017RobustPre} and our face-fairness. }	\label{fig:fair_plane_histograms}
\end{figure}

In the mesh denoising literature, a similar technique \cite{He2013MeshL0,Lu2017RobustPre} to avoid face-flipping is recently being used by defining a penalty across all the edges in the mesh. We however prefer to define our fairness term around the vertices. We observe that our face fairness term is more effective, resulting in better face shapes than defining the penalty on the edges as in \cite{Lu2017RobustPre}.
To demonstrate this fact, we perform the following experiment. We take a plane grid mesh of grid-size $36\times 36$. We then collapse some edges to deliberately introduce irregular and folded faces as shown in the first column in Figure \ref{fig:fair_plane}. 
With the assumption that we know the true normals of the surface, we apply our optimisation and also the method used in \cite{Lu2017RobustPre}. We find that our fairness penalty is better than \cite{Lu2017RobustPre} in two aspects. Firstly, our fairness results in better flat-shaped triangles whereas although \cite{Lu2017RobustPre} successfully unfolds the flipped faces, it results in more skinny triangles than there are in the initial irregular mesh. Secondly, in \cite{Lu2017RobustPre}, the effect of improving the triangles is not localised as shown in the second column in Figure \ref{fig:fair_plane}. To verify this fact quantitatively, we also compute the histograms of the face corner angles in the initial mesh and the two optimised meshes. As shown in Figure \ref{fig:fair_plane_histograms}, the histograms have two large peaks at $45^{\circ}$ and $90^{\circ}$, which is natural as these meshes arise from a grid. However, we note that the initial mesh has a profile (red coloured) showing a large fraction of the face corners has near zeros or very large ($>130^{\circ}$) angles. The output from the method of \cite{Lu2017RobustPre} has a profile (green coloured) that  is much flatter than the initial one showing that although the flipped faces are removed, the triangles become skinnier. However, output from our method has a profile (blue coloured) shows significant decrease in the number of face corners  with near-zero or very large angles.

\section{Applications} 

\subsection{Mesh denoising} \label{subsec:Denoising}

We now apply our framework to the problem of mesh denoising. Apart from being robust to discontinuities at the edges, we seek to account for the presence of noise in all directions by using our fairness framework which explicitly induces fairing of face shapes in the denoised mesh.\\

The operator $\mathbf{A}_{ij}$ in Equation \ref{eqn:vertex_modification_step} depends on the local geometry using the normals which is more sensitive to noise than the vertex positions themselves~\cite{Jones2003NonIterative}. Therefore, a mesh denoising algorithm requires a \textit{normal mollification} step for smoothing the normals before \textit{vertex correction}. Mollification is often implemented in two different ways~\cite{Sun2007FastEffective}, \textsl{i.e.}
\begin{enumerate}
	\item methods that iteratively improve the normals and vertex positions in an interleaved fashion, or 
	\item methods that first mollify the normals and then correct the vertex positions.
	
\end{enumerate}
In the latter case which move the noisy vertices only along the mollified normal directions, due to the large discrepancy between the mollified normals with respect to the noisy input vertices, the output mesh ends up with faces that are folded.\\

\subsubsection{Adaptation to denoising}

As said in the previous section, to adapt our fairness framework to the denoising 
 function, we first need to mollify the normals required to estimate the linear operators $\mathbf{A}_{ij}$. We pose our normal mollification as the minimisation of a global cost function. Being quadratic and sparse, our cost function has a solution which can be efficiently computed unlike methods that minimises the $\ell_{0}$-norm of certain functions. Once mollified normals are available, our fairness mesh optimisation framework can be directly applied. However, for proper denoising of 3D meshes, a pre-filtering is required \cite{Lu2017RobustPre}. So, we apply the above two steps twice. By explicitly incorporating the face fairness penalty term, our method removes noise both along the normal as well as in the tangent plane about a vertex. Such an approach coupled with a careful design of the data-adaptive weights of our cost function leads to proper denoising while ensuring fairness of the face shapes.\\

\subsubsection*{Mollification\label{sub:Mollification}}

The first step of our denoising approach is a mollification of the normals. As explained in the previous section, the local operators $\mathbf{A}_{ij}$ and weights $w_{ij}$ depend on local surface properties and need to be estimated from the noisy mesh $\mathbf{M}$. Hence, while the neighbourhood or topology of $\mathbf{M}$ is assumed to be unaltered, in our method we only require to mollify the set of face normals $\left\{ \mathbf{n}_{i}^{F}\right\} _{i=1}^{N_{F}}$. However unlike previous methods like \cite{Jones2003NonIterative,Sun2007FastEffective} we minimise a global cost function defined on the set of face normals, to obtain $\left\{ \widehat{\mathbf{n}}_{i}^{F}\right\} _{i=1}^{N_{F}}$ which is a smoothed version of the noisy face normals.\\

For each face normal $\mathbf{n}^{F}_{i}$, our cost function contains two terms, \textsl{i.e.} i) a data term $d_{o}^{N}\left(\widehat{\mathbf{n}}_{i}^{F},\mathbf{n}_{i}^{F}\right)$ which applies a quadratic penalty to the difference between the observed and estimated normals and ii) a weighted quadratic smoothness term over a local neighbourhood ${\displaystyle \sum_{j\in\mathcal{N}_F\left(i\right)}w_{ij}^{2}d_{s}^{N}\left(\widehat{\mathbf{n}}_{j}^{F},\widehat{\mathbf{n}}_{i}^{F}\right)}$ which induces local anisotropic smoothness. By adding up all the terms for each face normal, \textsl{i.e.} summing over index $i$, our solution for global mollification becomes one of minimising 

\begin{align}
&&{\displaystyle \sum_{i=1}^{N_{F}}d_{o}^{N}\left(\widehat{\mathbf{n}}_{i}^{F},\mathbf{n}_{i}^{F}\right)} +\lambda_{N}{\displaystyle \sum_{i=1}^{N_{F}}\sum_{j\in\mathcal{N}_{F}\left(i\right)}w_{ij}^{2}d_{s}^{N}\left(\widehat{\mathbf{n}}_{j}^{F},\widehat{\mathbf{n}}_{i}^{F}\right)}\nonumber \\
&&\textrm{ subject to }  {||\widehat{\mathbf{n}}_{i}^{F}||}^{2} = 1, \,i=1,2,\cdots , N_F\label{eq:mollification}
\end{align}

 where $\lambda_{N}$ is a regularising parameter depending on the noise variance and the face neighbourhood operator $\mathcal{N}_{F}\left(i\right)$ is defined as the set of faces which share a common vertex with the face $\mathbf{f}_{i}$ which is depicted in Fig.~\ref{fig:Notations}. We use the weighting function of \cite{Zheng2011Normal}, \textsl{i.e.}
\begin{eqnarray}
w_{ij}\left(\widehat{\mathbf{n}}_{j}^{F},\widehat{\mathbf{n}}_{i}^{F}\right)\nonumber
=
A_j\textrm{exp}\left( \frac{\left||\widehat{\mathbf{n}}_{j}^{F}-\widehat{\mathbf{n}}_{i}^{F}\right||^2}{2\sigma_{1}^2}+\frac{\left||\widehat{\mathbf{c}}_{j}-\widehat{\mathbf{c}}_{i}\right||^2}{2\sigma_{2}^2}\right)
 \label{eq:threshold}
\end{eqnarray}

where $A_j$ is the area of face $j$ and $\widehat{\mathbf{c}}_{i}$ and $\widehat{\mathbf{c}}_{j}$ are the centroids of faces $i$ and $j$.\\

The cost function in Equation~\ref{eq:mollification} is minimised using gradient descent. In the case where the noise level is moderate or high, we recompute weights $w_{ij}$ at every iteration of our gradient descent approach.\\

\subsubsection*{Vertex correction\label{sub:Vertex-Correction}}

Once we obtain an estimate of the face normals $\left\{ \widehat{\mathbf{n}}_{i}^{F}\right\} _{i=1}^{N_{F}}$, we can apply our vertex correction step. Most of the previous denoising methods update the vertex positions only in the normal direction. As we demonstrate in Section~\ref{sub:Synthetically-added-noisy}, for many existing methods this leads to irregular and folded faces in the denoised mesh which in turn results in shadow artefacts when we use smooth-shaded rendering mode in graphics pipelines.\\

\subsubsection{Results\label{sec:Results}}

We compare the denoising performance of our algorithm with some of the relevant methods in the literature. Specifically, we compared our results with the bilateral normal filtering \cite{Zheng2011Normal} and guided mesh denoising \cite{Zhang2015GuidedNormal}. \\

\subsubsection*{Quantitative evaluation of performance\label{sub:Synthetically-added-noisy}}

\begin{figure*}[h]
\noindent \begin{center}
\includegraphics[width=0.16\linewidth]{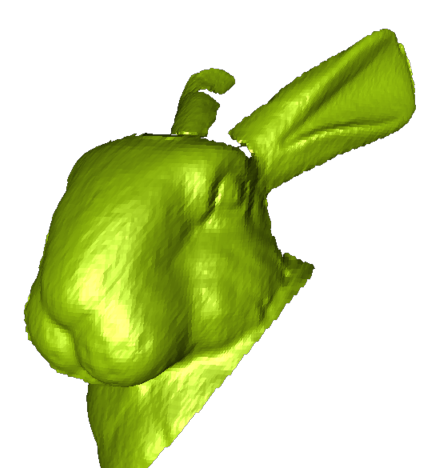}\includegraphics[width=0.16\linewidth]{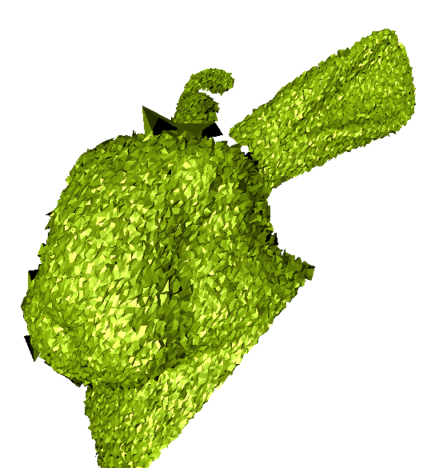}\includegraphics[width=0.16\linewidth]{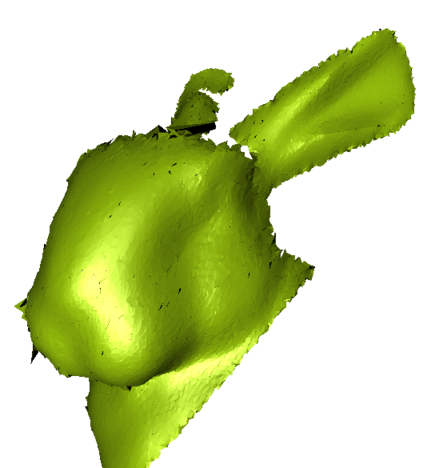}\includegraphics[width=0.16\linewidth]{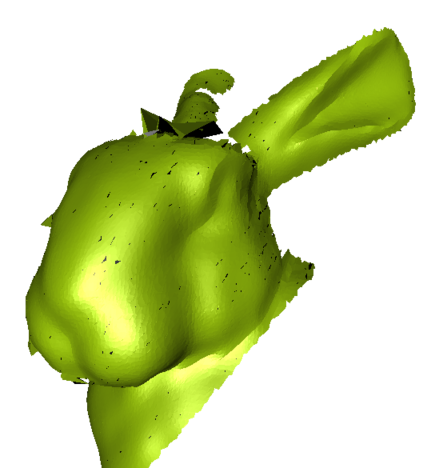}\includegraphics[width=0.16\linewidth]{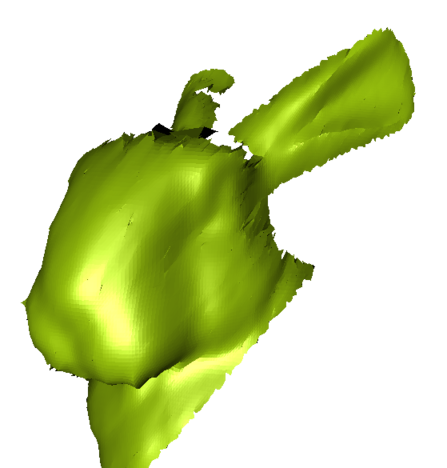}
\includegraphics[width=0.16\linewidth]{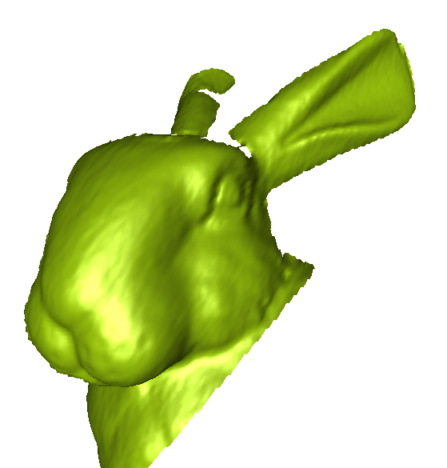}\includegraphics[width=0.16\linewidth]{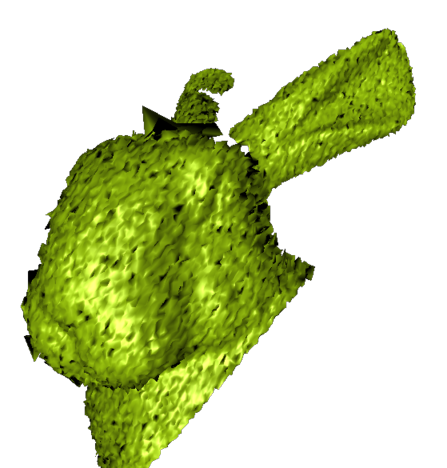}\includegraphics[width=0.16\linewidth]{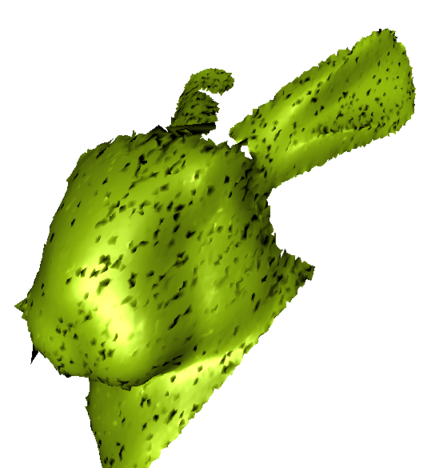}\includegraphics[width=0.16\linewidth]{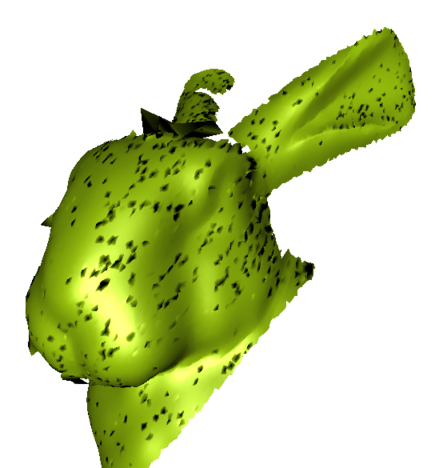}\includegraphics[width=0.16\linewidth]{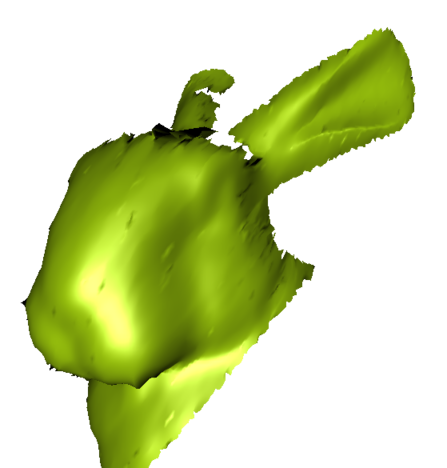}
\fnote{Ground truth\hspace{0.09\linewidth}Noisy\hspace{0.1\linewidth}\cite{Zheng2011Normal}\hspace{0.13\linewidth}\cite{Zhang2015GuidedNormal}\hspace{0.12\linewidth}Ours\hspace{0.06\linewidth}.}
\par\end{center}
  
\protect\caption{Denoised mesh quality of different methods on the bunny face ($N_V=11614,N_F=22574$) corrupted with isotropic Gaussian noise with standard deviation $\sigma = 0.35\times \textrm{mean edge length}$. The columns correspond to the ground truth, noisy mesh and solutions for bilateral normal filtering~\cite{Zheng2011Normal}, guided normal filtering~\cite{Zhang2015GuidedNormal} and our method respectively. The first row shows the surface quality. The second row shows the same surface in a smooth-shaded rendering mode. In this mode, the folded face artefacts prominently appear as black spots for the two methods other than ours.}\label{fig:Bunny_folded}
\end{figure*}

We first demonstrate the denoising performance of our method on three meshes, namely,  sphere, fandisk and bunny face, corrupted with additive Gaussian noise to the vertex positions as compared to some of the other relevant methods. For our evaluation, we use both the mean and median absolute deviation of the face normals measured in degrees and the mean and median Euclidean error metric on the vertex positions. We manually tune the parameters of each of the methods we used for best performance with respect to the mean face normal deviations. The comparisons are tabulated in Table~\ref{Table:Synthetic_evaluation}. It can be observed that, in all the cases, our method has the lowest vertex position errors. This is mainly due to the addition of the fairness penalty for the face shapes which carefully denoises the vertex positions on the surface along the tangential directions. Additionally, we can easily see that in all of the cases, the mean normal error is much higher in other two approaches compared to ours. This is due to the presence of folded faces in the outputs of these two approaches. However, the median normal error of guided mesh denoising~\cite{Zhang2015GuidedNormal} is lowest. In Fig.~\ref{fig:Bunny_folded} the first row shows the surface quality of the results obtained from different methods applied on the Bunny scan from the Stanford repository ($N_V=11614,N_F=22574$) corrupted with isotropic Gaussian noise with standard deviation $\sigma = 0.35\times \textrm{mean edge length}$. The second row shows the same results rendered in the smooth-shaded mode in OpenGL revealing the presence of folded faces as regions of black spots. Our result minimises such artefacts.\\

\begin{table*}
{\footnotesize
\noindent \begin{center}

\begin{tabular}{|cc|c|c|c|c|c|}
\hline 
 \multicolumn{2}{|c|}{Object} & Error metric & Noisy & BNF~\cite{Zheng2011Normal} & Guided~\cite{Zhang2015GuidedNormal} & Ours\tabularnewline
\hline 
\hline 
\multirow{4}{*}{\begin{minipage}{1.75in}Sphere\\($N_V=962,N_F=1920$)\\$\sigma = 0.35\times \textrm{mean edge length}$\end{minipage}}& \multirow{4}{*}{\includegraphics[scale=0.25]{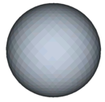}} & Mean NE ($^{\circ}$) & 32.0975 & 9.5190 & 4.5707 & \textbf{4.2478} \tabularnewline
\cline{3-7} 
  & & Median NE ($^{\circ}$) & 25.9236 & 3.0155 & \textbf{1.7523} & 2.6380 \tabularnewline
\cline{3-7}
  & & Mean VPE & 0.0306 & 0.0247 & 0.0137 & \textbf{0.0114} \tabularnewline
\cline{3-7}
 & & Median VPE & 0.0281 & 0.0221 & 0.0124 & \textbf{0.0100} \tabularnewline
\hline 
\multirow{4}{*}{\begin{minipage}{1.75in}Fandisk\\ ($N_V=6475,N_F=12946$)\\$\sigma = 0.35\times \textrm{mean edge length}$\end{minipage}} & \multirow{4}{*}{\includegraphics[scale=0.25]{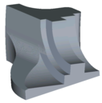}} & Mean NE ($^{\circ}$) & 31.4467 & 10.4579 & 5.4810 & \textbf{4.4946} \tabularnewline
\cline{3-7} 
  & & Median NE ($^{\circ}$) & 26.5515 & 2.3897 & \textbf{1.7121} & 1.9092 \tabularnewline
\cline{3-7} 
 & & Mean VPE & 0.0158 & 0.0149 & 0.0123 & \textbf{0.0080} \tabularnewline
\cline{3-7}
 & & Median VPE & 0.0150 & 0.0133 & 0.0111 & \textbf{0.0066} \tabularnewline
\hline 
\multirow{4}{*}{\begin{minipage}{1.75in}Bunny face\\($N_V=11614,N_F=22574$)\\$\sigma = 0.35\times \textrm{mean edge length}$ \end{minipage}} & \multirow{4}{*}{\includegraphics[scale=0.22]{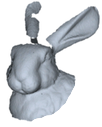}} & MeanNE ($^{\circ}$) & 38.6925 & 19.2335 & 15.1061 & \textbf{10.5908} \tabularnewline
\cline{3-7}
 & & Median NE ($^{\circ}$) & 30.2703 & 7.1014 & \textbf{5.1168} & 5.5742\tabularnewline
\cline{3-7}
 & & Mean VPE & 0.0517 & 0.04844 & 0.0433 & \textbf{0.0388}\tabularnewline
\cline{3-7}
 & & Median VPE & 0.0431 & 0.0405 & 0.0341 & \textbf{0.0232} \tabularnewline
\hline 
\end{tabular}
\par\end{center}
}
\protect\caption{Comparison of denoising performance of our approach with other methods in the literature. We compare both normal angle error (NE) and Vertex position Euclidean distance (VPE) error. `NA' denotes `Not Available'. The best performance for each dataset is indicated in bold. See text for details.}\label{Table:Synthetic_evaluation}
\end{table*}

\subsubsection*{Evaluation on real datasets\label{sub:Real-noisy-datasets}}

\begin{figure}[h]
\noindent \begin{center}
\includegraphics[width=0.20\linewidth]{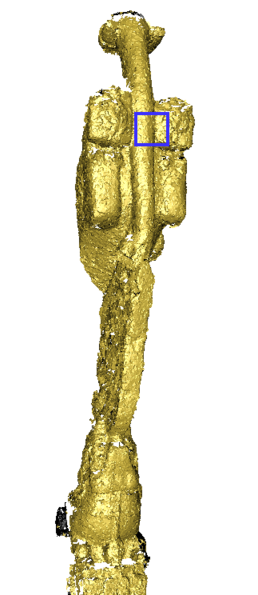}\,\includegraphics[width=0.20\linewidth]{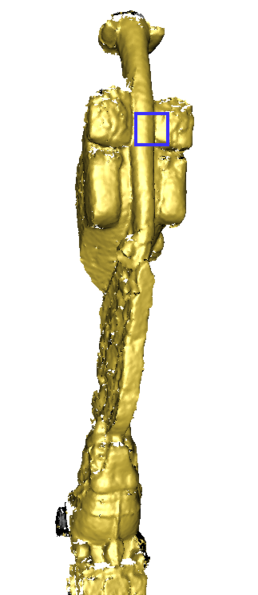}\,\includegraphics[width=0.20\linewidth]{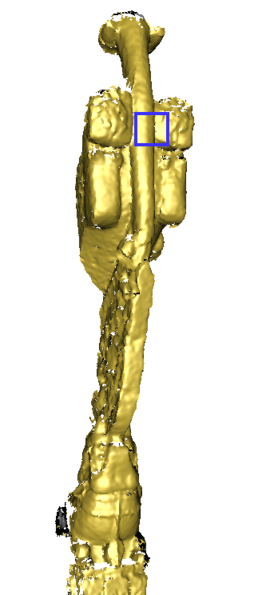}\,\includegraphics[width=0.20\linewidth]{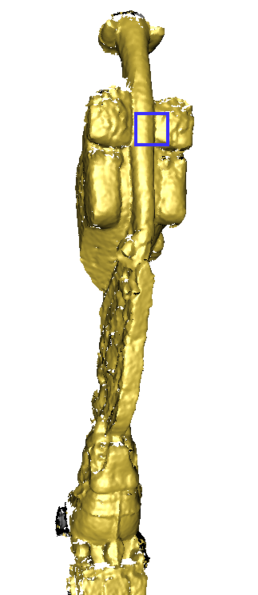}
\includegraphics[width=0.20\linewidth]{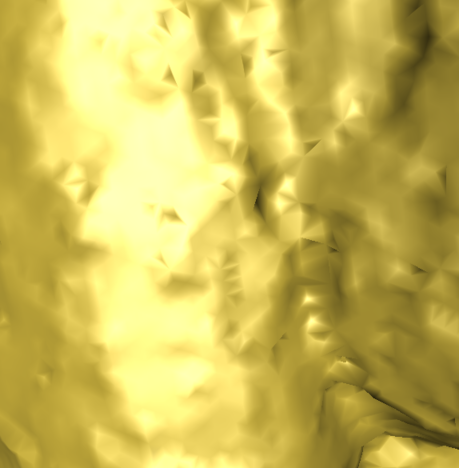}\,\includegraphics[width=0.20\linewidth]{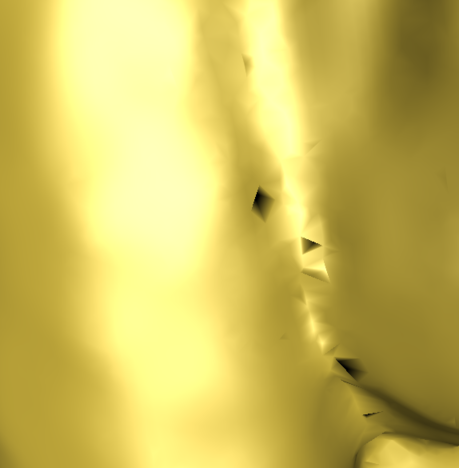}\,\includegraphics[width=0.20\linewidth]{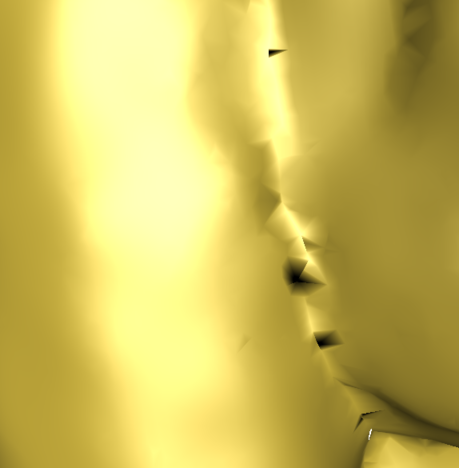}\,\includegraphics[width=0.20\linewidth]{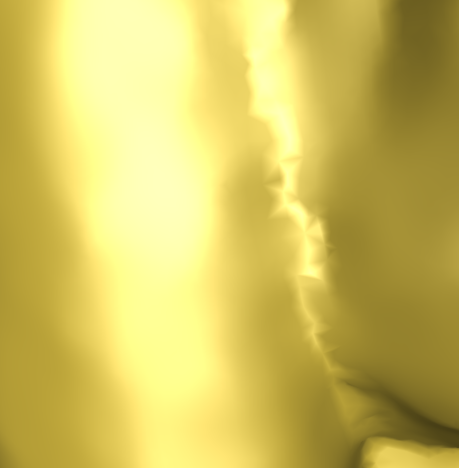}

\fnote{Noisy\hspace{0.12\linewidth}\cite{Zheng2011Normal}\hspace{0.13\linewidth}\cite{Zhang2015GuidedNormal}\hspace{0.13\linewidth}Ours}
\par\end{center}

\protect\caption{Denoising quality of different methods on a mesh ($N_V=185546,N_F=360814$) of a sculptural pillar generated using multi-view stereo applied to a set of images. The columns correspond to the noisy mesh and  bilateral normal filtering~\cite{Zheng2011Normal}, guided normal filtering~\cite{Zhang2015GuidedNormal} and our method respectively. The first row shows the overall surface. The second row shows the zoomed-in views with marked regions that show the presence of folded faces visible as black spots in smooth-shaded rendering mode. Such artefacts are not present in our denoised mesh.}\label{fig:Real_data_hampi}
\end{figure}

We present visual comparison on a real dataset. Fig.~\ref{fig:Real_data_hampi} shows the denoising results on a mesh ($N_V=185546$,$N_F=360814$) of a sculptured pillar from the Vitthala temple complex at Hampi, a heritage site. This mesh was generated using a standard multi-view stereo package on a set of RGB images of the pillar~\cite{FurukawaPMVS}. The second row shows the zoomed-in views (in smooth-shaded rendering mode) of the region marked as a black square. Our method  has the lowest number of folded mesh faces which is due to our explicit incorporation of a fairness penalty term.\\

\subsection{Mesh normal fusion}\label{subsec:mesh-normal-fusion}

Consumer-grade depth scanners are of low quality resulting in low quality 3D reconstructions. Although, the low frequency content of the depth estimates from
these scanners is known to be reliable, their high frequency content is rather poor \cite{Nehab2005Combine}. To improve the quality of such 3D reconstructions, a number of methods have been proposed. These methods use additional information like RGB colour images or IR images to complement the depth estimates with reliable high frequency details. These methods can be broadly classified into two groups, namely, a) implicit methods and b) explicit methods. While implicit methods are in general tuned to run in real-time \cite{El_2015_CVPR, Innmann2016Volumedeform}, the explicit methods generate higher quality 3D reconstructions offline.\\

In the explicit methods \cite{Haque2014High,Chatterjee2015Photometric},
a smooth low quality 3D mesh is generated from the raw depth maps
using volumetric methods like TSDF \cite{Curless1996Volumetric}.
The high frequency content is explicitly and independently recovered
as normal maps. These normal maps are then fused with the initial
smooth low quality 3D mesh to obtain the final reconstruction. One popular method that has been used in the literature is that of Nehab \textsl{et al.}~\cite{Nehab2005Combine, Wu2011Fusing, Haque2015Mesh, Haque2017MVRefinement}. In contrast to our proposed method, in the formulation of \cite{Nehab2005Combine}, the vertex positions are not allowed to move in the tangential directions to prevent face flipping. This is needed in their approach as their method does not enforce any explicit face-fairness penalty. The result, in turn, is the lowering of the degrees of freedom of the vertices to adequately fit to the high quality normals.
%We use the same smoothing framework which has a number of desirable properties unlike conventional methods that exists in the literature. As later, it is shown that such a framework leads to better quality results than the conventional methods.

\subsubsection{Adaptation to mesh normal fusion}

In the context of mesh-normal fusion, we have an initial smooth mesh $\mathbf{M}_{s}=\left(\mathbf{V}_{s},\mathbf{E},\mathbf{F}\right)$ that has low frequency fidelity and a normal map $\mathbf{N}_{d}$ containing high frequency details defined on the vertices $\mathbf{V}_{s}$ by their respective normals $\mathbf{N}_{h}^{V}$ over $\mathbf{M}_{s}$. We would then like the initial vertices $\mathbf{V}_s$ to locally fit these normals $\mathbf{N}_{h}^{V}$ or equivalently, we would like to ensure that the Laplacian of our desired mesh to be determined by the high quality input normal map $\mathbf{N}_{h}^{V}$. Simultaneously, we would like to make the final surface as smooth as possible. To achieve this, we use the vertex-normal map $\mathbf{N}_{h}^{V}$ to define $\mathbf{A}_{j}$ in Equation \ref{eqn:A_operator}. However, since $\mathbf{A}_{j}$ is defined over face normals, we use a weighted averaging scheme to transform the high quality vertex normal field $\mathbf{N}_{h}^{V}$ to a face normal field $\mathbf{N}_{h}^{F}$. For the $j^{th}$ face, we compute $\mathbf{n}_{h,j}^{F}$ as 
\begin{equation}
\mathbf{n}_{h,j}^{F} = \textrm{normalise}\left(\displaystyle\sum_{i\in \mathcal{N_F}(j)} w_{ij}\mathbf{n}_{h,i}^{N} \right)
\label{eqn:vertex_to_face_normals}
\end{equation}
where $w_{ij}=\max (\mathbf{n}_{h,i}^{V,T}\mathbf{n}_{s,j}^{F}, 0)$ and $\mathbf{n}_{s,j}^{F}$ is the normal on the $j^{th}$ face of the initial smooth mesh. Similarly, in the face fairness term in Equation \ref{eqn:fairness-function-per-vertex}, $\mathbf{n}_{i}^{V}$ is replaced with $\mathbf{n}_{h,}^{V}$ to use the high quality normals.

 As can be seen, the above transformation together with the Laplacian term in Equation \ref{eqn:laplacian} implicitly induces smoothness on the estimated surface, thereby minimising any residual noise in the normal map. Moreover, coupled with the face fairness term, our method allows more degrees of freedom to adequately fit the vertices to the high quality input normal field and simultaneously increasing the regularity of the face shapes.

\subsubsection{Results}

To evaluate the performance of our method, we consider the final mesh-normal fusion step of the 3D reconstruction pipeline proposed in \cite{Haque2014High,Haque2017MVRefinement}. We compare our method with the method of Nehab \textsl{et al.}~\cite{Nehab2005Combine} and a version of ours without the fairness term. We test on a number of real datasets.

\begin{figure*}[h]
\noindent \begin{centering}
\includegraphics[width=0.18\linewidth]{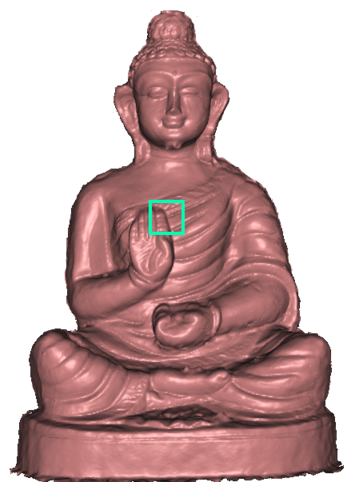}
\includegraphics[width=0.18\linewidth]{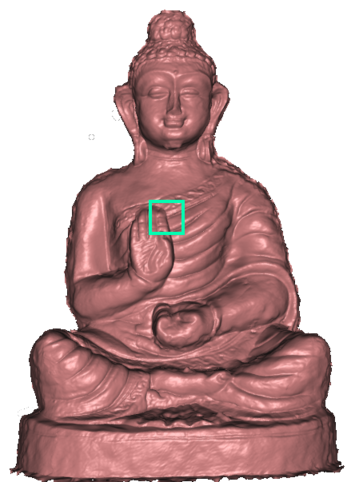}
\includegraphics[width=0.18\linewidth]{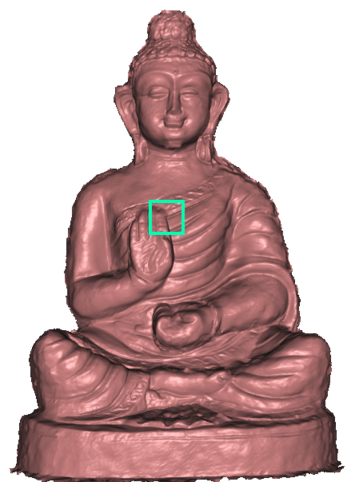}
\includegraphics[width=0.18\linewidth]{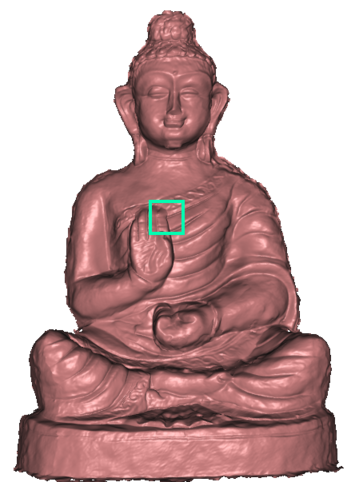}\\

\includegraphics[width=0.18\linewidth]{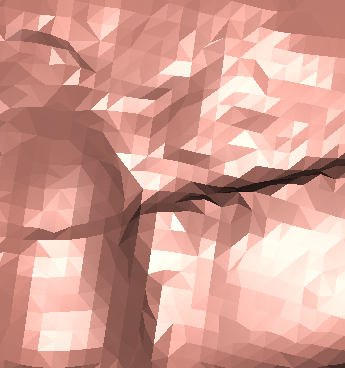}
\includegraphics[width=0.18\linewidth]{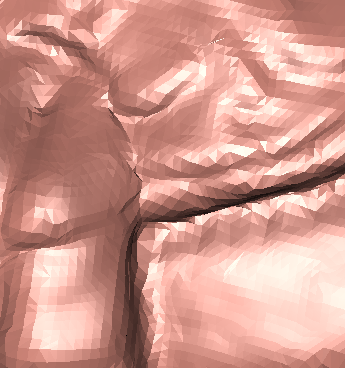}
\includegraphics[width=0.18\linewidth]{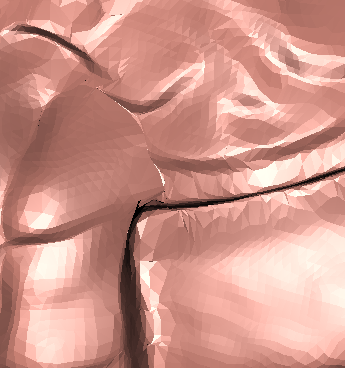}
\includegraphics[width=0.18\linewidth]{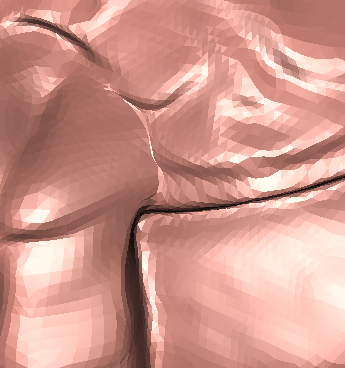}\\

\includegraphics[width=0.18\linewidth]{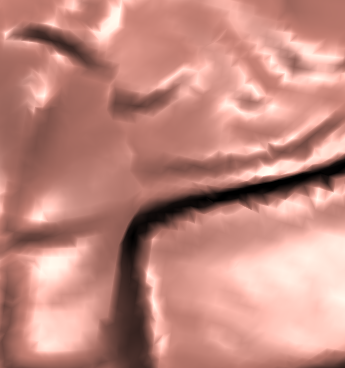}
\includegraphics[width=0.18\linewidth]{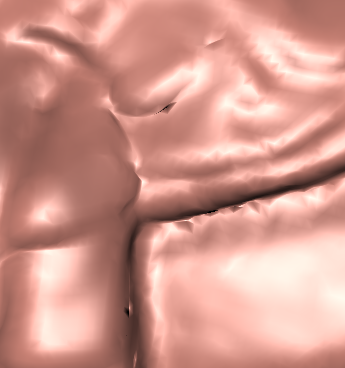}
\includegraphics[width=0.18\linewidth]{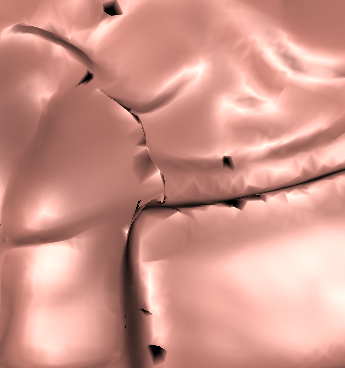}
\includegraphics[width=0.18\linewidth]{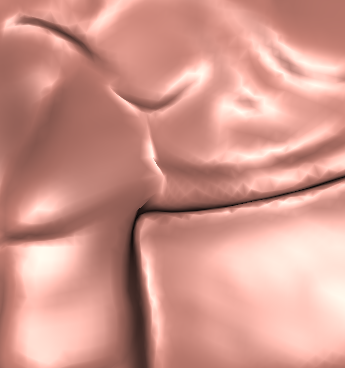}\\

\includegraphics[width=0.18\linewidth]{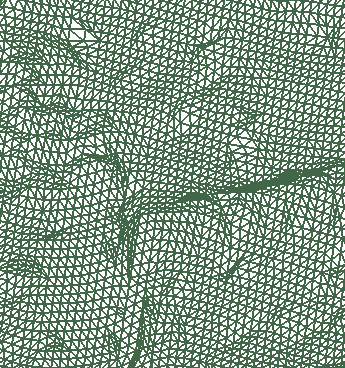}
\includegraphics[width=0.18\linewidth]{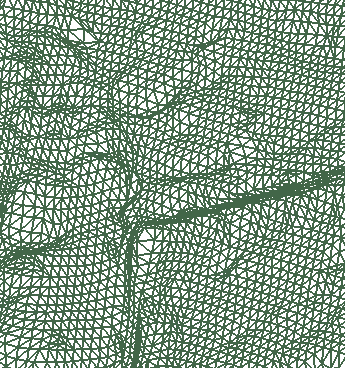}
\includegraphics[width=0.18\linewidth]{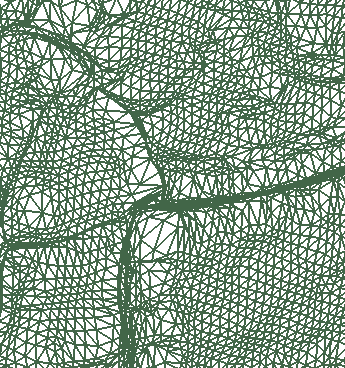}
\includegraphics[width=0.18\linewidth]{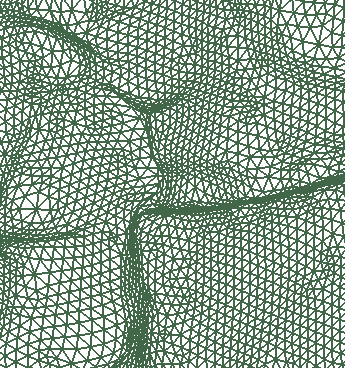}

\fnote{\hspace{0.05\linewidth}Initial\hspace{0.14\linewidth}\cite{Nehab2005Combine}\hspace{0.1\linewidth}Ours (without fairness)\hspace{0.035\linewidth}Ours (with fairness)}

\par\end{centering}

\protect\caption{Visual comparison on terracotta Buddha model. The rows correspond to the full refined meshes, zoomed-in views (flat-shaded and smooth-shaded) and the corresponding underlying mesh-edges of the highlighted regions respectively. The columns correspond to the initial mesh, refined meshes using \cite{Nehab2005Combine}, our method (without face fairness) and our method (with face fairness) respectively.}\label{fig:mnf_buddha}
\end{figure*}

Figure \ref{fig:mnf_buddha} shows a visual comparison of the results of applying the three different mesh refinement techniques using photometric normals on a terracotta Buddha model. The rows correspond to the full refined meshes, zoomed-in views (flat-shaded and smooth-shaded) and the corresponding underlying mesh-edges of the highlighted regions respectively. The columns correspond to the initial mesh, refined meshes using \cite{Nehab2005Combine}, our method (without face fairness) and our method (with face fairness) respectively. It can be observed from the second row that the output from Nehab \textsl{et al.}~\cite{Nehab2005Combine} has not adequately refined the mesh and is noisy at the edges, whereas in our method, both of our methods (with and without face fairness) has recovered smooth surface with sharp edges. However, from the third row, it can be seen that the addition of our fairness penalty has removed the flipped faces (black spots). Not only that, it can also be seen from the fourth row that our method has improved the shapes of the triangles at the same time.

\begin{figure*}[h]
\noindent \begin{centering}
\includegraphics[width=0.18\linewidth]{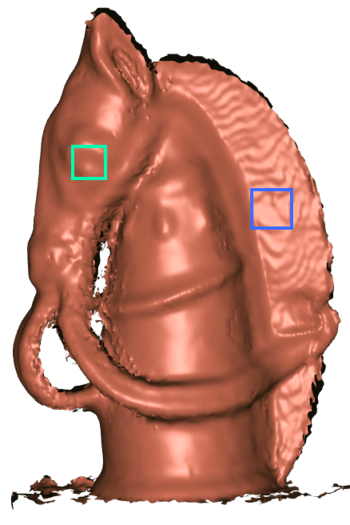}
\includegraphics[width=0.18\linewidth]{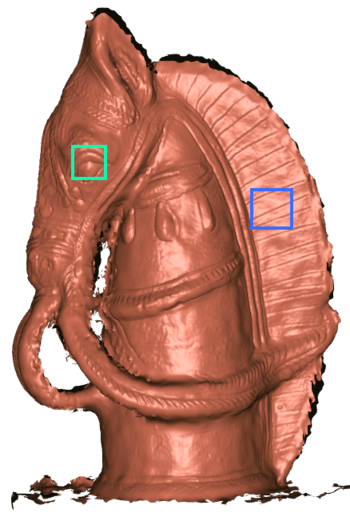}
\includegraphics[width=0.18\linewidth]{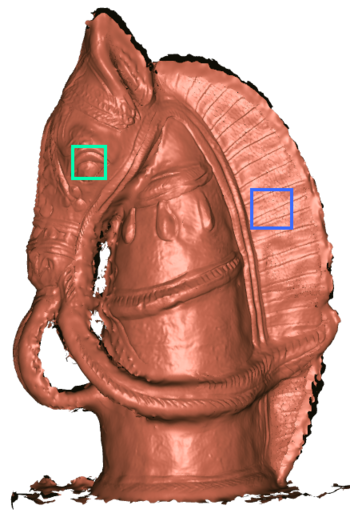}
\includegraphics[width=0.18\linewidth]{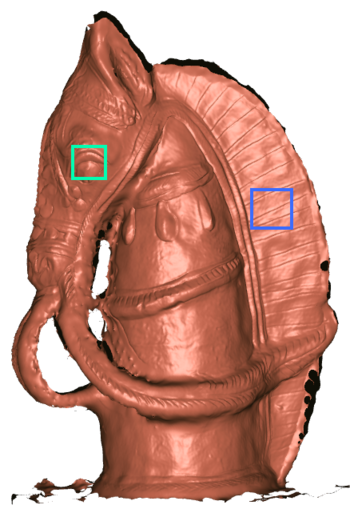}

\includegraphics[width=0.18\linewidth]{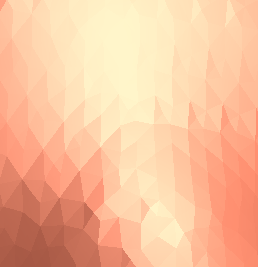}
\includegraphics[width=0.18\linewidth]{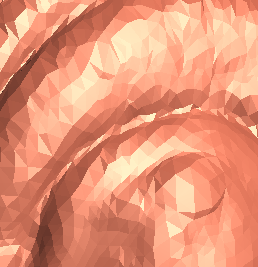}
\includegraphics[width=0.18\linewidth]{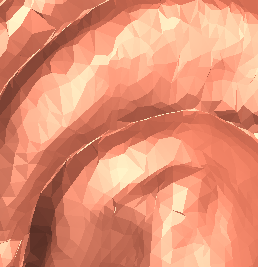}
\includegraphics[width=0.18\linewidth]{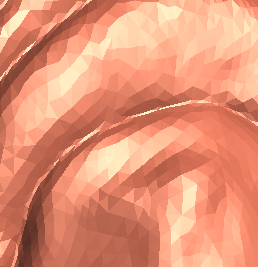}

\includegraphics[width=0.18\linewidth]{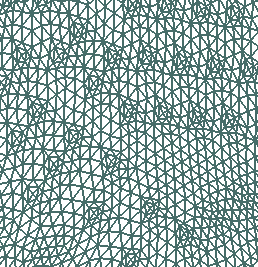}
\includegraphics[width=0.18\linewidth]{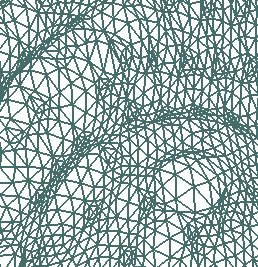}
\includegraphics[width=0.18\linewidth]{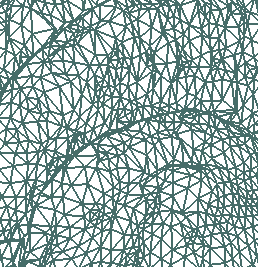}
\includegraphics[width=0.18\linewidth]{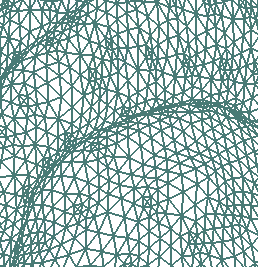}

\includegraphics[width=0.18\linewidth]{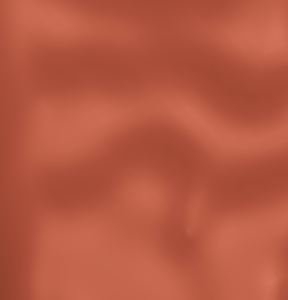}
\includegraphics[width=0.18\linewidth]{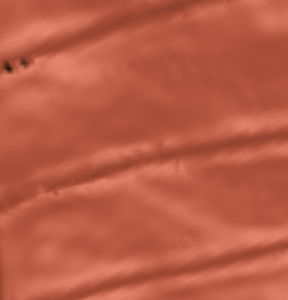}
\includegraphics[width=0.18\linewidth]{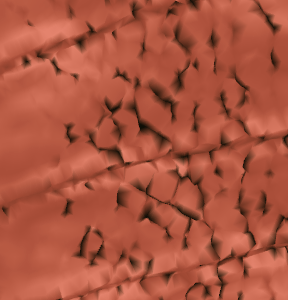}
\includegraphics[width=0.18\linewidth]{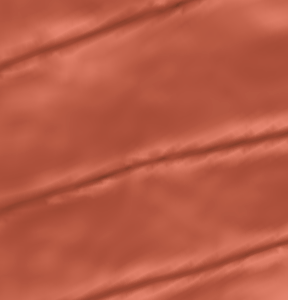}

\includegraphics[width=0.18\linewidth]{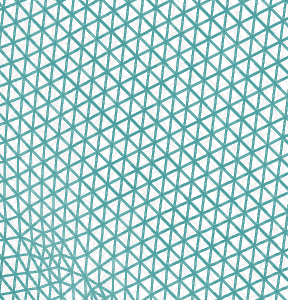}
\includegraphics[width=0.18\linewidth]{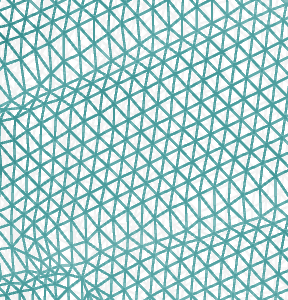}
\includegraphics[width=0.18\linewidth]{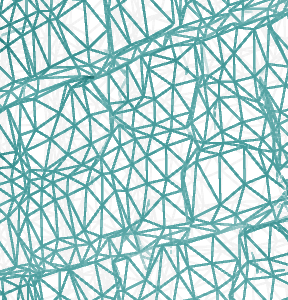}
\includegraphics[width=0.18\linewidth]{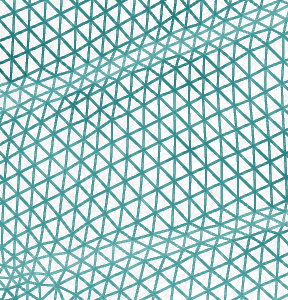}

\fnote{\hspace{0.05\linewidth}Initial\hspace{0.145\linewidth}\cite{Nehab2005Combine}\hspace{0.09\linewidth}Ours (without fairness)\hspace{0.016\linewidth}Ours (with fairness)}

\par\end{centering}

\protect\caption{Visual comparison on terracotta Horse model. The rows correspond to the full refined meshes, zoomed-in views and the corresponding underlying mesh-edges of the highlighted regions respectively (fourth row images are rendered in smooth-shaded mode.) The columns correspond to the initial mesh, refined meshes using \cite{Nehab2005Combine}, our method (without face fairness) and our method (with face fairness) respectively.}\label{fig:mnf_horse}
\end{figure*}

Figure \ref{fig:mnf_horse} shows a visual comparison of the results of applying the three different mesh refinement techniques using photometric normals on a terracotta Horse model. The second and third rows compare the eye region. Clearly, our method with face fairness (fourth column) has simultaneously recovered the fine edges and resulted in better-shaped triangles. The fourth and fifth rows compare the mane region of the horse. Although Nehab's method (second column) has preserved the triangle shapes, it has failed to recover as much detail of the edges as ours.

\begin{figure*}[h]
\noindent \begin{centering}
\includegraphics[width=0.2\linewidth]{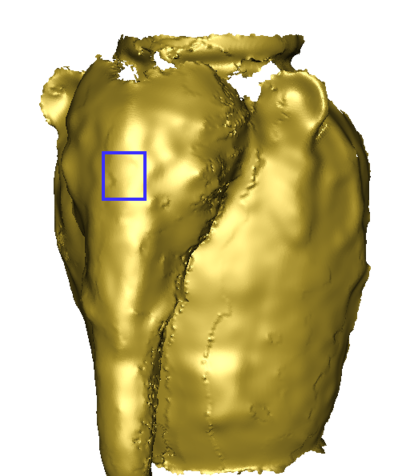}
\includegraphics[width=0.2\linewidth]{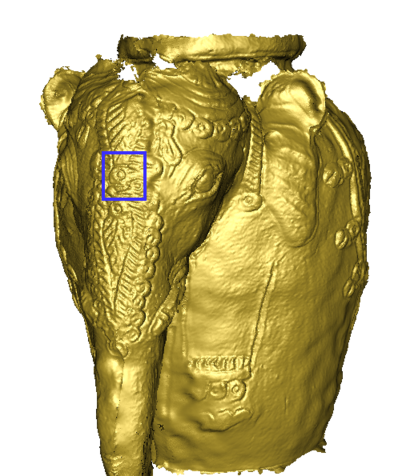}
\includegraphics[width=0.2\linewidth]{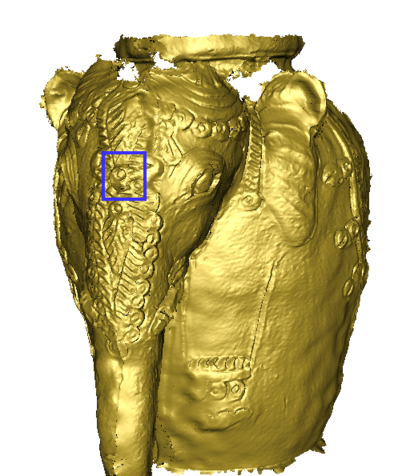}
\includegraphics[width=0.2\linewidth]{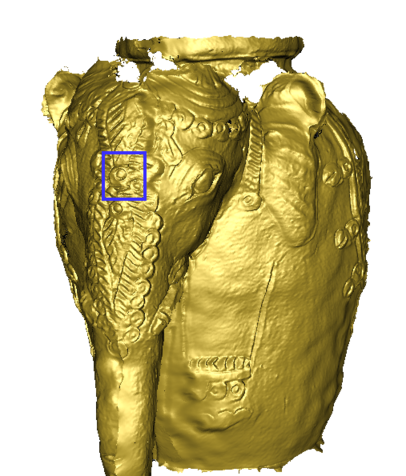}

\includegraphics[width=0.2\linewidth]{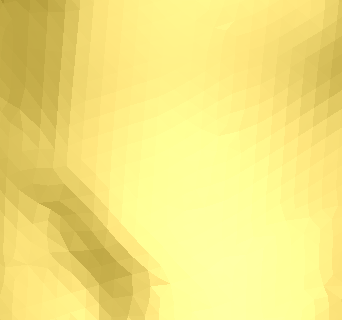}
\includegraphics[width=0.2\linewidth]{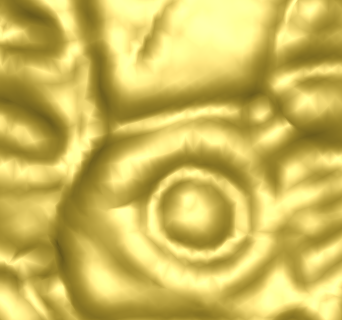}
\includegraphics[width=0.2\linewidth]{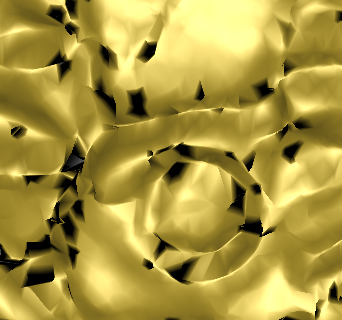}
\includegraphics[width=0.2\linewidth]{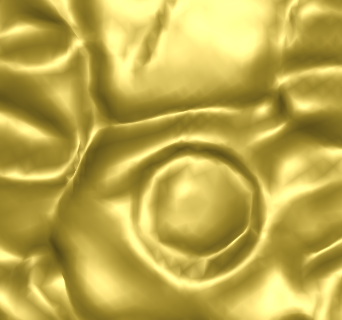}

\includegraphics[width=0.2\linewidth]{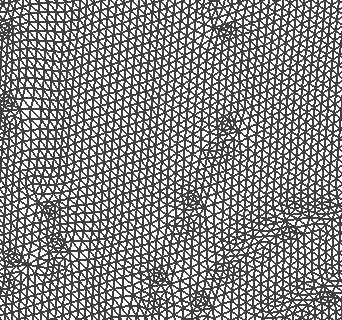}
\includegraphics[width=0.2\linewidth]{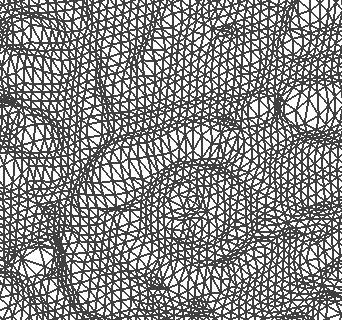}
\includegraphics[width=0.2\linewidth]{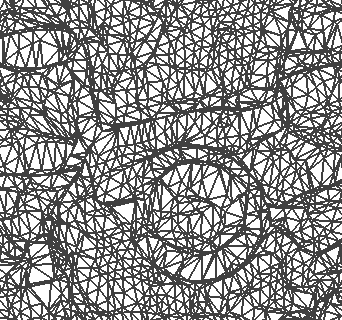}
\includegraphics[width=0.2\linewidth]{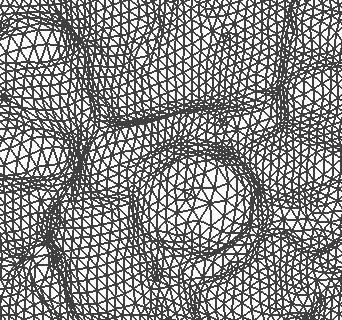}

\fnote{\hspace{0.05\linewidth}Initial\hspace{0.165\linewidth}\cite{Nehab2005Combine}\hspace{0.115\linewidth}Ours (without fairness)\hspace{0.05\linewidth}Ours (with fairness)}

\par\end{centering}

\protect\caption{Visual comparison on terracotta Jar model. The rows correspond to the full refined meshes, zoomed-in views (in smooth-shaded rendering mode) and the corresponding underlying mesh-edges of the highlighted regions respectively. The columns correspond to the initial mesh, refined meshes using \cite{Nehab2005Combine}, our method (without face fairness) and our method (with face fairness) respectively.}\label{fig:mnf_ele_jar}
\end{figure*}

Figure \ref{fig:mnf_ele_jar} shows a visual comparison of the results for a terracotta Jar model. As can be seen from the respective zoomed-in views of the fore-head region, our method with face fairness (fourth column) recovers sharp details and simultaneously preserves face shapes thus preventing occurrences of flipped faces in the refined meshes.

%
%\begin{figure}[H]
%	\noindent \begin{centering}
%		\textsf{\includegraphics[width=1\linewidth]{figures/fullhorse_a}}
%		\par\end{centering}
%	
%	\noindent \begin{centering}
%		\textsf{\includegraphics[width=1\linewidth]{figures/fullhorse_b}}
%		\par\end{centering}
%	
%	\protect\caption{Visual comparison on terracotta horse model}
%\end{figure}

\section{Conclusion\label{sec:Conclusion}}

We have presented a face fairness framework for 3D meshes that accounts for noise in all directions by incorporating a cost function for enforcing face fairness in the mesh. We have presented the applicability of our method to the tasks of mesh denoising and mesh-normal fusion. We have demonstrated the superiority of face fairness on a number of datasets.

\bibliographystyle{plain}
\bibliography{refs}

\begin{thebibliography}{10}

\bibitem{Babuvska1976Angle}
Ivo Babu{\v{s}}ka and A~Kadir Aziz.
\newblock On the angle condition in the finite element method.
\newblock {\em SIAM Journal on Numerical Analysis}, 13(2):214--226, 1976.

\bibitem{Chatterjee2015Photometric}
Avishek Chatterjee and Venu~Madhav Govindu.
\newblock Photometric refinement of depth maps for multi-albedo objects.
\newblock In {\em IEEE Conference on Computer Vision and Pattern Recognition},
  pages 933--941. IEEE, 2015.

\bibitem{Cheng2014FeatureL0}
Xuan Cheng, Ming Zeng, and Xinguo Liu.
\newblock Feature-preserving filtering with {$\ell_0$}-gradient minimization.
\newblock {\em Computers \& Graphics}, 38:150--157, 2014.

\bibitem{Curless1996Volumetric}
Brian Curless and Marc Levoy.
\newblock A volumetric method for building complex models from range images.
\newblock In {\em 23\textsuperscript{rd} Annual conference on Computer graphics
  and interactive techniques}, pages 303--312. ACM, 1996.

\bibitem{Desbrun1999Implicit}
Mathieu Desbrun, Mark Meyer, Peter Schr{\"o}der, and Alan~H Barr.
\newblock Implicit fairing of irregular meshes using diffusion and curvature
  flow.
\newblock In {\em 26\textsuperscript{th} Annual Conference on Computer Graphics
  and Interactive Techniques}, pages 317--324. ACM Press/Addison-Wesley
  Publishing Co., 1999.

\bibitem{Du1995Computing}
Ding-Zhu Du and Frank Hwang.
\newblock {\em Computing in Euclidean geometry}, volume~4.
\newblock World Scientific, 1995.

\bibitem{ElOuafdi2008PhysicalManifold}
A.F. El~Ouafdi and D.~Ziou.
\newblock A global physical method for manifold smoothing.
\newblock In {\em Shape Modeling and Applications, 2008. SMI 2008. IEEE
  International Conference on}, pages 11--17, June 2008.

\bibitem{Fan2010Robust}
Hanqi Fan, Yizhou Yu, and Qunsheng Peng.
\newblock Robust feature-preserving mesh denoising based on consistent
  subneighborhoods.
\newblock {\em IEEE Transactions on Visualization and Computer Graphics},
  16(2):312--324, 2010.

\bibitem{Field1988Laplacian}
David~A. Field.
\newblock Laplacian smoothing and delaunay triangulations.
\newblock {\em Communications in applied numerical methods}, 4(6):709--712,
  1988.

\bibitem{Fleishman2003Bilateral}
Shachar Fleishman, Iddo Drori, and Daniel Cohen-Or.
\newblock Bilateral mesh denoising.
\newblock {\em ACM Transactions on Graphics}, 22(3):950--953, 2003.

\bibitem{Fried1960Condition}
Isaac Fried.
\newblock Condition of finite element matrices generated from nonuniform
  meshes.
\newblock {\em Communications on Pure and Applied Mathematics}, 13:217--237,
  1960.

\bibitem{FurukawaPMVS}
Yasutaka Furukawa and Jean Ponce.
\newblock Accurate, dense, and robust multi-view stereopsis.
\newblock {\em IEEE Trans. on Pattern Analysis and Machine Intelligence},
  32(8):1362--1376, 2010.

\bibitem{Haque2014High}
Sk.~Mohammadul Haque, Avishek Chatterjee, and Venu~Madhav Govindu.
\newblock High quality photometric reconstruction using a depth camera.
\newblock In {\em IEEE Conference on Computer Vision and Pattern Recognition},
  pages 2275--2282, 2014.

\bibitem{Haque2015Mesh}
Sk.~Mohammadul Haque and Venu~Madhav Govindu.
\newblock Global mesh denoising with fairness.
\newblock In {\em 3rd International Conference on 3D Vision}, October 2015.

\bibitem{Haque2017MVRefinement}
Sk.~Mohammadul Haque and Venu~Madhav Govindu.
\newblock Multi-view non-rigid refinement and normal selection for high quality
  3d reconstruction.
\newblock In {\em IEEE International Conference on Computer Vision}, 2017.

\bibitem{He2013MeshL0}
Lei He and Scott Schaefer.
\newblock Mesh denoising via {$\ell_0$}-{m}inimization.
\newblock {\em ACM Transactions on Graphics}, 32(4):64, 2013.

\bibitem{Hoppe1993Optimization}
Hugues Hoppe, Tony DeRose, Tom Duchamp, John McDonald, and Werner Stuetzle.
\newblock Mesh optimization.
\newblock In {\em 20\textsuperscript{th} Annual conference on Computer graphics
  and interactive techniques}, pages 19--26. ACM, 1993.

\bibitem{Innmann2016Volumedeform}
Matthias Innmann, Michael Zollh{\"o}fer, Matthias Nie{\ss}ner, Christian
  Theobalt, and Marc Stamminger.
\newblock Volumedeform: Real-time volumetric non-rigid reconstruction.
\newblock In {\em European Conference on Computer Vision}, pages 362--379.
  Springer, 2016.

\bibitem{Ji2005Global}
Zhongping Ji, Ligang Liu, and Guojin Wang.
\newblock A global laplacian smoothing approach with feature preservation.
\newblock In {\em Ninth International Conference on Computer Aided Design and
  Computer Graphics}, pages 269--274. IEEE, IEEE Computer Society, 2005.

\bibitem{Jones2003NonIterative}
Thouis~R. Jones, Fr{\'e}do Durand, and Mathieu Desbrun.
\newblock Non-iterative, feature-preserving mesh smoothing.
\newblock {\em ACM Transactions on Graphics}, 22(3):943--949, July 2003.

\bibitem{Kim2016MVIR}
Kichang Kim, Akihiko Torii, and Masatoshi Okutomi.
\newblock Multi-view inverse rendering under arbitrary illumination and albedo.
\newblock In {\em European Conference on Computer Vision}, pages 750--767.
  Springer International Publishing, 2016.

\bibitem{Liu2007NonIterative}
Ligang Liu, Chiew-Lan Tai, Zhongping Ji, and Guojin Wang.
\newblock Non-iterative approach for global mesh optimization.
\newblock {\em Computer-Aided Design}, 39(9):772--782, 2007.

\bibitem{Lu2017RobustPre}
Xuequan Lu, Wenzhi Chen, and Scott Schaefer.
\newblock Robust mesh denoising via vertex pre-filtering and l1-median normal
  filtering.
\newblock {\em Computer Aided Geometric Design}, 2017.

\bibitem{Lu2017Efficient}
Xuequan Lu, Xiaohong Liu, Zhigang Deng, and Wenzhi Chen.
\newblock An efficient approach for feature-preserving mesh denoising.
\newblock {\em Optics and Lasers in Engineering}, 90:186--195, 2017.

\bibitem{Max1999Weights}
Nelson Max.
\newblock Weights for computing vertex normals from facet normals.
\newblock {\em Journal of Graphics Tools}, 4(2):1--6, 1999.

\bibitem{Meyer2003Discrete}
Mark Meyer, Mathieu Desbrun, Peter Schr{\"o}der, and Alan~H Barr.
\newblock Discrete differential-geometry operators for triangulated
  2-manifolds.
\newblock In {\em Visualization and mathematics III}, pages 35--57. Springer,
  2003.

\bibitem{Nealen2006Laplacian}
Andrew Nealen, Takeo Igarashi, Olga Sorkine, and Marc Alexa.
\newblock Laplacian mesh optimization.
\newblock In {\em 4\textsuperscript{th} international conference on Computer
  graphics and interactive techniques in Australasia and Southeast Asia}, pages
  381--389. ACM, 2006.

\bibitem{Nehab2005Combine}
D.~Nehab, S.~Rusinkiewicz, J.~E. Davis, and R.~Ramamoorthi.
\newblock Efficiently combining positions and normals for precise {3D}
  geometry.
\newblock {\em ACM Transactions on Graphics}, 24(3):536--543, 2005.

\bibitem{Ohtake02MeshSmoothing}
Yutaka Ohtake, Alexander Belyaev, and Hans{-}peter Seidel.
\newblock Mesh smoothing by adaptive and anisotropic gaussian filter applied to
  mesh normals.
\newblock In {\em Vision Modeling and Visualization}, 2002.

\bibitem{El_2015_CVPR}
Roy Or~El, Guy Rosman, Aaron Wetzler, Ron Kimmel, and Alfred~M. Bruckstein.
\newblock Rgbd-fusion: Real-time high precision depth recovery.
\newblock In {\em IEEE Conference on Computer Vision and Pattern Recognition},
  June 2015.

\bibitem{Sun2008RandomWalk}
Xianfang Sun, Paul~L. Rosin, Ralph~R. Martin, and Frank~C. Langbein.
\newblock Random walks for feature-preserving mesh denoising.
\newblock {\em Computer Aided Geometric Design}, 25(7):437--456, 2008.

\bibitem{Sun2007FastEffective}
Xianfang Sun, P.L. Rosin, R.R. Martin, and F.C. Langbein.
\newblock Fast and effective feature-preserving mesh denoising.
\newblock {\em IEEE Transactions on Visualization and Computer Graphics},
  13(5):925--938, September 2007.

\bibitem{Taubin1995Signal}
Gabriel Taubin.
\newblock A signal processing approach to fair surface design.
\newblock In {\em 22\textsuperscript{nd} Annual Conference on Computer Graphics
  and Interactive Techniques}, SIGGRAPH '95, pages 351--358, New York, NY, USA,
  1995. ACM.

\bibitem{Taubin2001LinearAniso}
Gabriel Taubin.
\newblock Linear anisotropic mesh filtering.
\newblock {\em Res. Rep. RC2213 IBM}, 2001.

\bibitem{Thurrner1998Computing}
Grit Th{\"u}rrner and Charles~A W{\"u}thrich.
\newblock Computing vertex normals from polygonal facets.
\newblock {\em Journal of Graphics Tools}, 3(1):43--46, 1998.

\bibitem{Wang2014DNF}
Ruimin Wang, Zhouwang Yang, Ligang Liu, Jiansong Deng, and Falai Chen.
\newblock Decoupling noise and features via weighted {$\ell_1$}-analysis
  compressed sensing.
\newblock {\em ACM Transactions on Graphics}, 33(2):18:1--18:12, April 2014.

\bibitem{Wei2015BiNormal}
Mingqiang Wei, Jinze Yu, Wai-Man Pang, Jun Wang, Jing Qin, Ligang Liu, and
  Pheng-Ann Heng.
\newblock Bi-normal filtering for mesh denoising.
\newblock {\em IEEE Transactions on Visualization and Computer Graphics},
  21(1):43--55, January 2015.

\bibitem{Wu2011HQMultiviewStereo}
C.~Wu, B.~Wilburn, Y.~Matsushita, and C.~Theobalt.
\newblock High-quality shape from multi-view stereo and shading under general
  illumination.
\newblock In {\em IEEE Conference on Computer Vision and Pattern Recognition},
  pages 969--976, June 2011.

\bibitem{Wu2011Fusing}
Chenglei Wu, Yebin Liu, Qionghai Dai, and Bennett Wilburn.
\newblock Fusing multiview and photometric stereo for 3d reconstruction under
  uncalibrated illumination.
\newblock {\em IEEE Transactions on Visualization and Computer Graphics},
  17(8):1082--1095, 2011.

\bibitem{Zhang2015Variational}
Huayan Zhang, Chunlin Wu, Juyong Zhang, and Jiansong Deng.
\newblock Variational mesh denoising using total variation and piecewise
  constant function space.
\newblock {\em IEEE Transactions on Visualization and Computer Graphics},
  21(7):873--886, July 2015.

\bibitem{Zhang2015GuidedNormal}
Wangyu Zhang, Bailin Deng, Juyong Zhang, Sofien Bouaziz, and Ligang Liu.
\newblock Guided mesh normal filtering.
\newblock {\em Computer Graphics Forum}, 34(7):23--34, 2015.

\bibitem{Zheng2011Normal}
Youyi Zheng, Hongbo Fu, O.K.-C. Au, and Chiew-Lan Tai.
\newblock Bilateral normal filtering for mesh denoising.
\newblock {\em IEEE Transactions on Visualization and Computer Graphics},
  17(10):1521--1530, October 2011.

\end{thebibliography}

\end{document}